\documentclass{tlp}

\usepackage{times}
\usepackage[T1]{fontenc}
\usepackage{amsmath,amssymb}
\usepackage{epsfig}

\usepackage[dvips]{color}

\newcommand{\wrt}{{\it w.r.t. }}   
\newcommand{\eg}{\emph{e.g.}}       
\newcommand{\ie}{\emph{i.e.}}      
\newcommand{\etal}{\emph{et al.}}         
\newcommand\etc{\emph{etc.}}

\newtheorem{corollary}{Corollary}

\newtheorem{proposition}{Proposition}
\newtheorem{example}{Example}
\newtheorem{definition}{Definition}
\newtheorem{remark}{Remark}

\newenvironment{statement}[1][Statement]{\vspace*{0.5cm} \noindent  \textbf{#1} \it }

\submitted{October 10, 2008}
 \revised{March 22, 2011}
 \accepted{May 23, 2011}

\begin{document}

\title[Semantics for Possibilistic Disjunctive Programs]
{Semantics for Possibilistic Disjunctive Programs \thanks{This is a
revised and improved version of the papers
\emph{Semantics for
Possibilistic Disjunctive Programs} appeared in C. Baral,
G. Brewka and J. Schipf (Eds), Ninth International Conference on
Logic Programming and Nonmonotonic Reasoning (LPNMR-07), LNAI 4483.
\emph{Semantics for Possibilistic Disjunctive Logic programs} which
appears in S. Constantini and W. Watson (Eds), Answer Set
Programming: Advantage in Theory and Implementation.}}

\author[J. C. Nieves, M. Osorio, and U. Cort\'es]
{ JUAN CARLOS NIEVES\\
Universitat Polit\`ecnica de Catalunya\\
Software Department (LSI)\\
c/Jordi Girona 1-3, E-08034, Barcelona, Spain  \\
\email{jcnieves@lsi.upc.edu} \and
MAURICIO OSORIO \\
Universidad de las Am\'ericas - Puebla\\
CENTIA \\
Sta. Catarina M\'artir, Cholula, Puebla, 72820 M\'exico\\
\email{osoriomauri@gmail.com}
\and
ULISES CORT\'ES  \\
Universitat Polit\`ecnica de Catalunya\\
Software Department (LSI)\\
c/Jordi Girona 1-3, E-08034, Barcelona, Spain  \\
\email{ia@lsi.upc.edu}
}

\maketitle

\begin{abstract}
In this paper, a possibilistic disjunctive logic programming approach for
modeling uncertain, incomplete and inconsistent information is defined. This approach introduces the use of possibilistic disjunctive clauses
which are able to capture incomplete information and incomplete
states of a knowledge base at the same time.

By considering a possibilistic logic program as a possibilistic logic theory, a construction of a possibilistic logic programming semantic based on answer sets and the proof theory of possibilistic logic is defined. It shows that this possibilistic semantics for disjunctive logic programs can be characterized by a fixed-point operator. It is also shown that the suggested possibilistic semantics can be computed by a resolution algorithm and the consideration of optimal refutations from a possibilistic logic theory.

In order to manage inconsistent possibilistic logic programs, a
preference criterion between inconsistent possibilistic models is
defined; in addition, the approach of cuts for restoring consistency
of an inconsistent possibilistic knowledge base is adopted. The
approach is illustrated in a medical scenario.

\end{abstract}

\begin{keywords}
Answer Set Programming, Uncertain Information, Possibilistic
Reasoning.
\end{keywords}

\section{Introduction}\label{sec:PossProg-Motivation}

Answer Set Programming (ASP) is one of the most successful logic
programming approaches in Non-monotonic Reasoning and Artificial
Intelligence applications \cite{Bar03,Gel08}. In
\cite{NicGarSteLef06}, a possibilistic framework for reasoning under
uncertainty was proposed. This framework is a combination between
ASP and possibilistic logic \cite{Dubois94}.

Possibilistic Logic is based on possibilistic theory in which, at the
mathematical level, degrees of possibility and necessity are closely
related to fuzzy sets \cite{Dubois94}. Due to the natural
properties of possibilistic logic and ASP, Nicolas \etal's approach
allows us to deal with reasoning that is at the same time
\emph{non-monotonic} and \emph{uncertain}. Nicolas \etal's approach
is based on the concept of \emph{possibilistic stable model} which
defines a semantics for \emph{possibilistic normal logic programs}.

An important property of possibilistic logic is that it is \emph{axiomatizable} in the necessity-valued case \cite{Dubois94}. This means that there is a formal system (a set of axioms and inferences rules) such that from any set of possibilistic fomul\ae\ ${\cal F}$ and for any possibilistic formula $\Phi$, $\Phi$ is a logical consequence of ${\cal F}$ if and only if $\Phi$ is derivable from ${\cal F}$ in this formal system.
A result of this property is that the inference in possibilistic logic can be managed by both a syntactic approach  (axioms and inference rules) and a possibilistic model theory approach (interpretations and possibilistic distributions).

Equally important to consider is that
the answer set semantics inference can also be characterized as \emph{a logic inference} in terms of the proof theory of intuitionistic logic and intermediate logics \cite{Pea99,OsNaAr:tplp}. This property suggests that one can explore extensions of the answer set semantics by considering the inference of different logics.

Since in \cite{Dubois94} an axiomatization  of possibilistic logic has been defined,
in this paper we explore the characterization of a possibilistic semantics for capturing possibilistic logic programs in terms of the proof theory of possibilistic logic and the standard answer set semantics. A nice feature of this characterization is that it is applicable to \emph{disjunctive as well as normal possibilistic logic programs}, and, with minor modification, to possibilistic logic programs containing a \emph{strong negation operator}.

%
%

The use of possibilistic disjunctive logic programs allow us to capture \emph{incomplete information} and \emph{incomplete states of a knowledge base} at the same time. In order to illustrate the use of possibilistic disjunctive logic programs, let us consider a scenario in which uncertain and incomplete information is always present. This scenario can be observed in the process of \emph{human organ transplanting}. There are several factors that make this process sophisticated and complex. For instance:

\begin{itemize}
  \item the transplant acceptance criteria vary ostensibly among
transplant teams from the same geographical area and substantially
between more distant transplant teams \cite{LopDomVie97}. This
means that the acceptance criteria applied in one hospital could be
invalid or at least questionable in another hospital.

  \item
there are lots of factors that make the diagnosis of an organ donor's disease in the organ recipient unpredictable.
For instance, if an organ  donor $D$ has hepatitis, then an organ
recipient $R$ could be infected by an organ of $D$. According to
\cite{LopCab03}, there are cases in which the infection can occur;
however, the recipient can \emph{spontaneously} clear the infection, for example hepatitis.
This means that an organ donor's infection can be
\emph{present} or \emph{non-present} in the organ recipient.
Of course there are infections which can be prevented by treating the organ recipient post-transplant.


\item the clinical state of an organ recipient can be affected by
several factors, for example malfunctions of the graft. This means that the
clinical state of an organ recipient can be \emph{stable} or
\emph{unstable} after the graft because the graft can have
\emph{good graft functions}, \emph{delayed graft functions} and
\emph{terminal insufficient functions}\footnote{Usually, when a
doctor says that an organ has \emph{terminally insufficient
functions}, it  means that there are no clinical treatments for
improving the organ's functions.}.
\end{itemize}

It is important to point out that the transplant acceptance criteria rely on
the kind of organ (kidney, heart, liver, \etc)
considered for transplant and the clinical situation of
the potential organ recipients.

\begin{figure}[h] 
\begin{center}
\epsfig{file=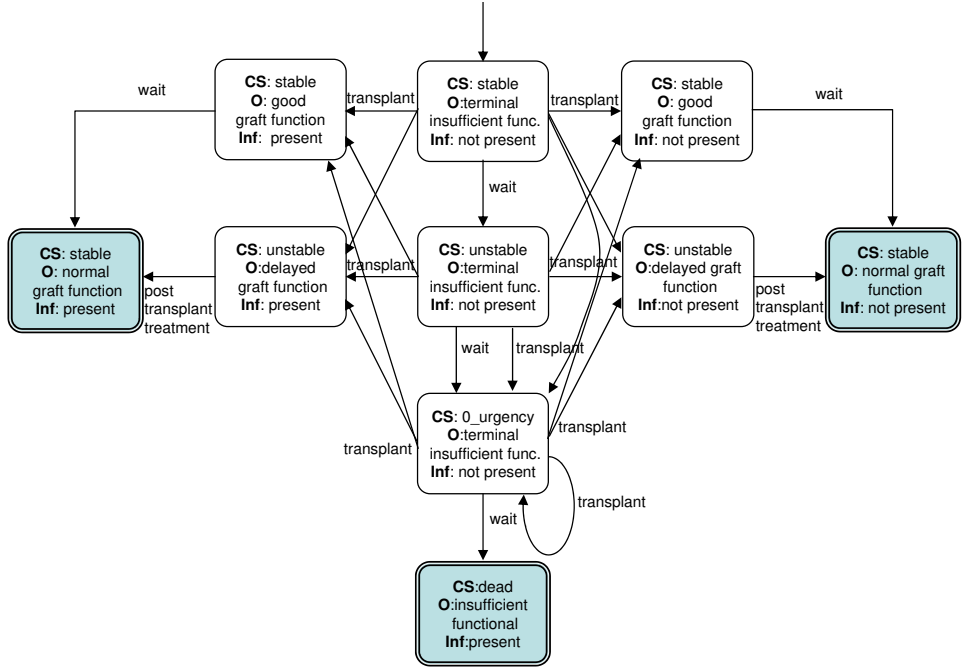,scale=0.65,angle=270, bbllx=86,
bblly=133, bburx=475, bbury=703, clip=} \caption{An automata of
states and actions for considering infections in kidney organ
transplant.}\label{automata-actions-infections}
\end{center}
\end{figure}

Let us consider the particular case of a kind of kidney transplant with
organ donors who have a kind of infection,  for example:
endocarditis, hepatitis. As already stated, the clinical situation of the
potential organ recipients is relevant in the organ transplant
process. Hence the clinical situation of an organ recipient is denoted
by the predicate $cs(t, T)$, such that $t$ can be \emph{stable},
\emph{unstable}, 0-\emph{urgency} and $T$ denotes a moment in
time. Another important factor, that is considered, is the state of
the organ's functions. This factor is denoted by the predicate $o(t, T)$
such that $t$ can be \emph{terminal-insufficient functions}, \emph{good-graft functions},
\emph{delayed-graft functions}, \emph{normal-graft functions} and
$T$ denotes a moment in time. Also, the state of an infection in both the organ recipient and
the organ donor are considered, these states are denoted by the predicates
$r\_inf(present,T)$ and $d\_inf(present,T)$ respectively so that
$T$ denotes a moment in time. The last predicate that is presented is $action(t,T)$ such that $t$ can be
\emph{transplant}, \emph{wait}, \emph{post-transplant treatment} and
$T$ denotes a moment in time. This predicate denotes the
possible actions of a doctor. In Figure
\ref{automata-actions-infections}\footnote{This finite state
automata was developed under the supervision of Francisco Caballero M.
D. Ph. D. from the Hospital de la Santa Creu I Sant Pau, Barcelona,
Spain.}, a finite state automata is presented.
In this automata, each node represents a possible situation where an organ
recipient can be found and the arrows represent the doctor's possible
actions. Observe that we are assuming that in the initial
state the organ recipient is clinically stable and he does not have
an infection; however, he has a kidney whose functions are terminally
insufficient. From the initial state, the doctor's actions would be
either to perform a kidney transplantat or just wait\footnote{In the
automata of Figure \ref{automata-actions-infections}, we are not
considering the possibility that there is a waiting list for organs. This waiting list has different policies for
assigning an organ to an organ recipient.}.


According to Figure \ref{automata-actions-infections}, an organ recipient could  be found in different situations after a graft. The organ recipient may require another graft and the state of the infection could be unpredictable. This situation makes the automata of Figure \ref{automata-actions-infections} nondeterministic.
Let us consider a couple of extended disjunctive clauses which describe some situations presented in
Figure \ref{automata-actions-infections}. \\

\noindent
\begin{tabular}{l}
$r\_inf(present,T2) \vee \lnot r\_inf(present,T2)$ $\leftarrow$
$action(transplant,T),$ \\
$~~~~~~~~~~~~~~~~~~~~~~~~~~~~~~~~~ d\_inf(present,T), T2 = T +
1.$\\
\\
$o(good\_graft\_funct, T2) \vee
o(delayed\_graft\_funct, T2)\vee$ \\
$o(terminal\_insufficient\_funct, T2) \leftarrow
 action(transplant, T), T2 = T + 1. $ \\
\end{tabular}\\

\noindent

As syntactic clarification, we want to point out that $\lnot$ is
regarded as a \emph{strong negation} which is not exactly the negation
in classical logic. In fact, any atom negated by strong negation
will be replaced by a new atom as it is done in ASP. This means that
$a \vee \lnot a$ cannot be regarded as a logic tautology.

Continuing with our medical scenario, we can see that the intended
meaning of the first clause is that if the organ donor has an
infection, then the infection can be \emph{present} or
\emph{non-present} in the organ recipient after the graft, and the
intended meaning of the second one is that the graft's functions can
be: \emph{good}, \emph{delayed} and \emph{terminal} after the
graft. Observe that these clauses are not capturing the uncertainty
that is involved in each statement. For instance, \wrt the first
clause, one can wish to attach an degree of uncertainty in order to
capture the uncertainty that is involved in this statement ---
keeping in mind that the organ recipient can be infected by the infection
of the donor's organ; however, the infection can be
\emph{spontaneously} cleared by the organ recipient as it is the
case of hepatitis \cite{LopCab03}.

In logic programming literature, one can find different approaches for representing uncertain information  \cite{KifSub92,NgS92,Lukasiewicz98,Kern-IsbernerL04,Emden86,RodRome08,NieuwenborghCV07,Fitting91,Lakshmanan94,Bal07,DubLanPra01,AlsinetG02,AlsGod00,AlsinetCGS08,NicGarSteLef06}.
Basically, these approaches differ in the underlying notion of uncertainty and how uncertainty values, associated with clauses and facts, are managed. Usually the selection of an approach for representing uncertain information relies on the kind of information which has to be represented. In psychology literature, one can find significant observations related to the presentation of uncertain information. For instance, Tversky and Kahneman have observed in \cite{TveKan82} that people commonly use statements such as ``\emph{I think that} $\dots$'', ``\emph{chances are }$\dots$'', ``\emph{it is probable that} $\dots$'', ``\emph{it is plausible that} $\dots$'', \etc, for supporting their decisions. In fact, many times, experts in a domain, such as medicine, appeal to their intuition by using these kinds of statements \cite{FoxDas00,FoxSan06}.
One can observe that these statements have adjectives which quantify the information as a common denominator.
These adjectives are for example: \emph{probable}, \emph{plausible}, \emph{etc}. This suggests that the consideration of labels for the \emph{syntactic representation of uncertain values }could help  represent uncertain information pervaded by ambiguity.

Since possibilistic logic defines a proof theory in which the strength of a conclusion is the strength of the weakest argument in its proof, the consideration of an ordered set of labels for capturing incomplete states of a knowledge base is feasible.
The only formal requirement is that this set of adjectives/labels must
be a finite set.
For instance, for the given medical scenario, a transplant coordinator\footnote{A transplant
coordinator is an expert in all of the processes of transplants \cite{LopDomVie97}.} can suggest a set of labels in order to quantify a medical knowledge base and, of course, to define an order between those labels. By considering those labels, we can have possibilistic clauses as:\\


\noindent  \textbf{probable}: $r\_inf(present, T2) \vee \lnot
r\_inf(present,T2)$ $\leftarrow$ $action(transplant, T),$ \\
$~~~~~~~~~~~~~~~~~~~~~~~~~~~~~~~~~~~~ d\_inf(present, T), T2 = T + 1.$ \\


\noindent Informally speaking, the reading of this clause is:
\emph{it is \emph{probable} that if the organ donor has an infection, then  the organ recipient can be infected or not after a graft.}

As we can see, possibilistic programs with \emph{negation as failure} represent a rich class of logic programs which are especially adapted to automated reasoning when the available information is pervaded by ambiguity.

In this paper, we extend the work of two earlier papers \cite{NivOsoCor07b,NivOsoCorASP07} in order to obtain a simple logic characterization of a possibilistic logic programming semantics for capturing possibilistic programs; this semantics is applicable to disjunctive as well as normal logic programs.
As we have already mentioned, the construction of the possibilistic semantics is based on the proof theory of possibilistic logic. Following this approach:

\begin{itemize}
  \item We define the inference $\Vvdash_{PL}$. This inference takes as references the standard definition of the answer set semantics and the inference $\vdash_{PL}$ which corresponds to the inference of possibilistic logic.
  \item The possibilistic semantics is defined in terms of a syntactic reduction, $\Vvdash_{PL}$ and the concept of \emph{i-greatest set}.
  \item Since the inference of possibilistic logic is computable by a generalization of the classical resolution rule, it is shown that the defined possibilistic semantics is computable by inferring optimal refutations.
  \item By considering \emph{the principle of partial evaluation}, it is shown that the given possibilistic semantics can be characterized by a possibilistic partial evaluation operator.
  \item Finally, since the possibilistic logic uses $\alpha$\emph{-cuts}  to manage inconsistent possibilistic knowledge bases, an approach of cuts for restoring consistency of an inconsistent possibilistic knowledge base is adopted.
\end{itemize}


The rest of the paper is divided as follows: In \S \ref{chap:back} we give all the background and necessary notation. In \S \ref{sec:Poss-syntax}, the syntax of our possibilistic framework is presented. In \S \ref{sec:Poss-Sems}, the semantics for capturing the possibilistic logic programs is defined. Also it is shown that this semantics is computable by considering a possibilistic resolution rule and partial evaluation. In \S \ref{sec:inconsitency}, some criteria for managing inconsistent possibilistic logic programs are defined.
In \S \ref{sec:relatedWorkPossDisjunctivePrograms}, we present a
small discussion \wrt related approaches to our work. Finally, in
the last section, we present our conclusions and future work.

\section{Background}\label{chap:back}

In this section we introduce the necessary terminology and
relevant definitions in order to have a self-contained document. We
assume that the reader is familiar with basic concepts of
\emph{classic logic}, \emph{logic programming} and \emph{lattices}.

\subsection{Lattices and order}


We start by defining some fundamental definitions of
lattice theory 
(see \cite{DavPri02} for more details).

\begin{definition}
Let ${\cal Q}$ be a set. An order (or partial order) on ${\cal Q}$
is a binary relation $\leq$ on ${\cal Q}$ such that, for all $x,y,z
\in {\cal Q}$,

\begin{description}
  \item[(i)]  $x \leq x$
  \item[(ii)] $x \leq y$ and $y \leq x$ imply $x = y$
  \item[(iii)] $x \leq y$ and $y \leq z$ imply $x \leq z$
\end{description}

\noindent These conditions are referred to, respectively, as
reflexivity, antisymmetry and transitivity.
\end{definition}

A set ${\cal Q}$ equipped with an order relation $\leq$ is said to
be an ordered set (or partial ordered set). It will be denoted by
(${\cal Q}$,$\leq$).

\begin{definition}
Let (${\cal Q}$,$\leq$) be an ordered set and let $S \subseteq {\cal
Q}$. An element $x \in {\cal Q}$ is an upper bound of $S$ if $s \leq
x$ for all $s \in S$. A lower bound is defined dually. The set of
all upper bounds of $S$ is denoted by $S^u$ (read as `$S$ upper')
and the set of all lower bounds by $S^l$ (read as `$S$ lower').
\end{definition}

If $S^u$ has a minimum element $x$, then $x$ is called the least upper
bound (${\cal LUB}$) of $S$. Equivalently, $x$ is the least upper
bound of $S$ if

\begin{description}
  \item[(i)] $x$ is an upper bound of $S$, and
  \item[(ii)] $x \leq y$ for all upper bound $y$ of $S$.
\end{description}

The least upper bound of $S$ exists if and only if there exists $x
\in {\cal Q}$ such that
$$(\forall y \in {\cal Q}) [ ((\forall s \in
S) s \leq y ) \Longleftrightarrow x \leq y ], $$

\noindent and this characterizes the ${\cal LUB}$ of $S$. Dually, if
$S^l$ has a greatest element, $x$, then $x$ is called the greatest
lower bound (${\cal GLB}$) of $S$. Since the least element and the greatest
element are unique, ${\cal LUB}$ and ${\cal GLB}$ are unique when
they exist.

The least upper bound of $S$ is called the supremum of $S$ and it is
denoted by $sup ~ S$; the greatest lower bound of S is called
the infimum of S and it is denoted by $inf ~ S.$

\begin{definition}
Let (${\cal Q}$,$\leq$) be a non-empty ordered set.

\begin{description}
  \item[(i)] If $sup \{x,y\}$ and $inf \{x,y\}$ exist for all $x,y \in
  {\cal Q}$, then ${\cal Q}$ is called lattice.
  \item[(ii)] If $sup ~S$ and $inf ~S$ exist for all $S \subseteq {\cal Q}$, then ${\cal Q}$
  is called a complete lattice.
\end{description}
\end{definition}

\begin{example}\label{ex:lattice-labels}
Let us consider the set of labels ${\cal Q} := \{Certain$,
$Confirmed,$ $Probable,$ $Plausible,$ $Supported,$ $Open\}$\footnote{This
set of labels was taken from \cite{FoxSan06}. In that paper, the
authors argue  that we can construct a set of labels (they call
those: \emph{modalities}) in a way that this set provides a simple
scale
for ordering the claims of our beliefs. 
We will use this kind of labels for quantifying the degree of uncertainty
of a statement.} and let $\preceq$ be a partial order such
that the following set of relations holds: $\{Open \preceq
Supported$, $Supported \preceq Plausible $, $Supported \preceq
Probable$, $Probable \preceq Confirmed$, $Plausible \preceq
Confirmed$, $Confirmed \preceq Certain\}$. A graphic representation
of $S$ according to $\preceq$ is showed in Figure
\ref{fig:lattice-background}. It is not difficult to see that
$({\cal Q}, \preceq)$ is a lattice and further it is a complete
lattice.

\begin{figure}[th]
\begin{center}
\epsfig{file=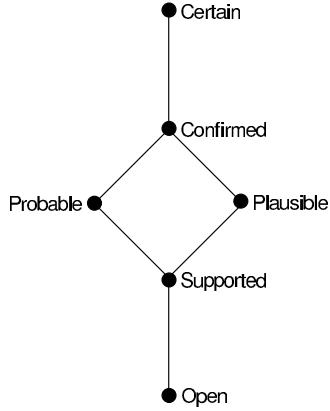,scale=0.6, bbllx=197, bblly=227,
bburx=420, bbury=500, clip=} \caption{A graphic representation of a
lattice where the following relations holds: $\{Open   \preceq
Supported$, $Supported \preceq Plausible $ $Supported \preceq
Probable$, $Probable \preceq Confirmed$, $Plausible \preceq
Confirmed$ , $Confirmed \preceq
Certain\}$.}\label{fig:lattice-background}
\end{center}
\end{figure}

\end{example}

\subsection{Logic programs: Syntax}\label{aspSyntaxis}

The language of a propositional logic has an alphabet consisting of
\begin{description}
\item[(i)] proposition symbols: $\bot, p_0, p_1, ...$
\item[(ii)] connectives : $\vee , \wedge , \leftarrow , \lnot, \; not$
\item[(iii)] auxiliary symbols : ( , )
\end{description}
in which $\vee , \wedge , \leftarrow$ are binary-place connectives,
$\lnot$, $not$ are unary-place connective and $\bot$ is zero-ary
connective. The proposition symbols and $\bot$  stand for the
indecomposable propositions, which we call {\it atoms}, or {\it
atomic propositions}. Atoms negated by $\lnot$ will be called
\emph{extended atoms}.

\begin{remark}
We will use the concept of atom without paying attention to whether it is an extended atom or not.
\end{remark}

\noindent The negation sign $\lnot$ is regarded as the so called
\emph{strong negation} by the ASP's literature and the negation
$not$ as the \emph{negation as failure}. A literal is an atom, $a$,
or the negation of an atom $not\; a$. Given a set of atoms
$\{a_{1},...,a_{n}\}$, we write $not \; \{a_{1},...,a_{n}\}$ to
denote the set of literals $\{not \; a_{1},..., not \; a_{n}\}.$ An
extended disjunctive clause, {\it C}, is denoted:
$$a_1 \vee \ldots \vee a_m \leftarrow  a_{m+1}, \dots, a_{j}, not \; a_{j+1}, \dots, not\; a_n
$$

\noindent in which $m \geq 0$, $n \geq 0$, $m + n > 0$, each $a_i$ is
an atom\footnote{Notice that these atoms can be \emph{extended
atoms}.}. When $n=0$ and $m>0$ the clause is an abbreviation of
$a_1\vee \ldots \vee a_m \leftarrow$;
clauses of these forms are some times written just as $a_1\vee \ldots \vee a_m$. When
$m=0$ the clause is an abbreviation of: $$\leftarrow a_1, \dots,
a_{j}, not \; a_{j+1}, \dots, not\; a_n$$
Clauses of this form are called constraints (the rest, non-constraint clauses). An
extended disjunctive program $P$ is a finite set of extended
disjunctive clauses. By ${\cal L}_P$, we denote the set of atoms in
the language of $P$.

Sometimes we denote an extended disjunctive clause {\it C} by $
{\cal A} \leftarrow {\cal B}^+ , \ not \; {\cal B}^-$, ${\cal
A}$ contains all the head literals, ${\cal B}^+$ contains all the
positive body literals and ${\cal B}^-$ contains all the negative
body literals. When ${\cal B}^- = \emptyset$, the clause is called
positive disjunctive clause. A set of positive disjunctive clauses
is called a positive disjunctive logic program. When ${\cal A}$ is a
singleton set, the clause can be regarded as a normal clause. A
normal logic program is a finite set of normal clauses. Finally,
when ${\cal A}$ is a singleton set and ${\cal B}^- = \emptyset$, the
clause can also be regarded as a definite clause. A finite set of
definite clauses is called a definite logic program.

We will manage the strong negation ($\neg$), in our logic programs,
as it is done in ASP \cite{Bar03}. Basically, each
extended atom $\neg a$ is replaced by a new atom symbol $a'$ which does not
appear in the language of the program.
For instance, let $P$ be the normal program: \\

\begin{tabular}{lll}
$a \leftarrow q$. & $~~~~~~~~~$ &$q$. \\
$\lnot q \leftarrow r$. & $~~~~~~~~~$ &$r$.\\
\end{tabular}\\

\noindent Then replacing each extended atom by a new atom symbol, we
will have: \\

\begin{tabular}{lll}
$a \leftarrow q$. & $~~~~~~~~~$ & $q$.\\
$q' \leftarrow r$. & $~~~~~~~~~$ & $r$.\\
\end{tabular} \\

In order not to allow models with complementary atoms, that is $q$ and
$\lnot q$, a constraint of the form $\leftarrow q, q'$ is usually added to the logic program. In our approach, this constraint can be omitted
in order to allow models with complementary atoms. In fact, the user could add/omit  this constraint without losing generality.

Formul\ae\ are constructed as usual in classic logic by the connectives: $\vee , \wedge , \leftarrow ,
\sim, \bot$. A theory $T$ is a finite set of formul\ae. By ${\cal L}_T$, we denote the set of atoms that occur in \emph{T}. When we treat a logic program as a theory,

\begin{itemize}
  \item each negative literal $not \; a$ is replaced by $\sim a$ such that $\sim$ is regarded as the negation in classic logic.
  \item each constraint $\leftarrow a_1, \dots, a_{j}, not \; a_{j+1}, \dots, not\; a_n$ is rewritten according to the formula $a_1 \wedge \dots \wedge a_{j} \wedge \sim  a_{j+1} \wedge \dots \wedge \sim  a_n \rightarrow \bot$.
\end{itemize}

Given a set of proposition symbols $S$ and a theory  $\Gamma$ in a logic $X$. If $\Gamma \vdash_{X} S$ if and only if $\forall s \in S$ $\Gamma \vdash_{X} s$.

\subsection{Interpretations and models}

In this section, we define some relevant concepts \wrt semantics. The
first basic concept that we introduce is \emph{interpretation}.

\begin{definition} Let $T$ be a theory, an interpretation $I$ is a mapping from
${\cal L}_T$ to $\{0, 1\}$ meeting the conditions:
\begin{enumerate}
\item $I(a \wedge b) = min\{I(a), I(b)\}$,
\item $I(a \vee b) = max\{I(a), I(b)\}$,
\item $I(a\leftarrow b) = 0$ if and only if $I(b) = 1$ and $I(a) = 0$,
\item $I(\sim a) = 1 - I(a)$,
\item $I(\bot) = 0$.
\end{enumerate}
\end{definition}

It is standard to provide interpretations only in terms of a mapping
from ${\cal L}_T$ to $\{0, 1\}$. Moreover, it is easy to prove that
this mapping is unique by virtue of the definition by recursion
\cite{Dalen94}. Also, it is standard to use sets of atoms to
represent interpretations. The set corresponds exactly to those
atoms that evaluate to 1.

An interpretation $I$ is called a (2-valued) model of the logic
program $P$ if and only if for each clause $c \in P$, $I(c) = 1$. A
theory is consistent if it admits a model, otherwise it is called
inconsistent. Given a theory $T$ and a formula $\varphi$, we say that
$\varphi$ is a logical consequence of $T$, denoted by $T \models
\varphi$, if every model $I$ of $T$ holds that $I(\varphi)=1$. It is a
well known result that $T \models \varphi$ if and only if $T \cup
\{\sim \varphi \}$ is inconsistent \cite{Dalen94}.

We say that a model $I$ of a theory $T$  is a minimal model  if
a model $I'$ of $T$ different from $I$ such that $I' \subset I$ does not exist.
Maximal models are defined in the analogous form.

\subsection{Logic programming semantics}

In this section, the \emph{answer set semantics} is presented. This semantics represents a two-valued semantics approach.

\subsubsection{Answer set semantics}\label{sec:ASP-semantics}

By using ASP, it is possible to describe a computational problem as
a logic program whose answer sets correspond to the solutions of the
given problem. It represents one of the most successful approaches
of non-monotonic reasoning of the last two decades \cite{Bar03}. The
number of applications of this approach have increased due to
the efficient implementations of the answer set solvers that exist.

The answer set semantics was first defined in terms of the so called
\emph{Gelfond-Lifschitz reduction} \cite{GelLif88} and it is usually
studied in the context of syntax dependent transformations on
programs. The following definition of an answer set for extended
disjunctive logic programs generalizes the definition presented in
\cite{GelLif88} and it was presented in \cite{GelLif91}: Let
\emph{P} be any extended disjunctive logic program. For any set $S
\subseteq {\cal L}_P$, let $P^S$ be the positive program obtained
from \emph{P} by deleting

\begin{description}
\item[(i)] each rule that has a formula $not \; a$ in its body with
$a \in S$, and then

\item[(ii)] all formul\ae\ of the form $not \; a$ in the bodies of
the remaining rules.
\end{description}

\noindent Clearly $P^S$ does not contain $not$ (this means that
$P^S$ is either a positive disjunctive logic program or a definite
logic program), hence \emph{S} is called an answer set of \emph{P}
if and only if \emph{S} is a minimal model of $P^S$. In order to
illustrate this definition, let us consider the following example:

\begin{example}
Let us consider the set of atoms  $S := \{ b \}$ and the following
normal logic program $P$:

\begin{tabular}{ll}
  $b \leftarrow not \; a$. $~~~~~~~~~~~~~~~~~$ & $b$. \\
  $c \leftarrow not \; b$. $~~~~~~~~~~~~~~~~~$ & $c \leftarrow a$. \\
\end{tabular}

\noindent We can see that $P^S$ is:

\begin{tabular}{ll}
  $b$. $~~~~~~~~~~~~~~~~~$ & $c \leftarrow a$. \\
\end{tabular}

\noindent Notice that this program has three models: $\{b\}$,
$\{b,c\}$ and $\{a, b,c\}$. Since the minimal model among these
models is $\{b\}$, we can say that $S$ is an answer set of $P$.
\end{example}

In the answer set definition, we will normally omit the restriction
that if $S$ has a pair of complementary  literals then $S := {\cal
L}_P$. This means that we allow for the possibility that an answer set could have a pair
of complementary atoms. For instance, let us consider the program
$P$:

\begin{tabular}{lllll}
$a$. & $~~~~~~~$ & $\lnot a$. & $~~~~~~~$ & $b$.
\end{tabular}

\noindent then, the only answer set of this program is : $\{a,\lnot
a, b\}$. In Section \ref{sec:inconsitency}, the inconsistency in possibilistic programs is discussed.


It is worth mentioning that in literature there are several forms
for handling an inconsistency program \cite{Bar03}. For instance, by applying the
original definition \cite{GelLif91} the only answer set of $P$ is:
$\{a, \lnot a, b, \lnot b\}$. On the other hand, the DLV system
\cite{DLV} returns no models if the program is inconsistent.

\subsection{Possibilistic Logic}\label{subsect:PS}

Since in our approach is based on the proof theory of possibilistic logic, in this section,
we present an axiomation  of possibilistic logic for the case of necessity-valued formul\ae.\

Possibilistic logic is a weighted logic introduced and developed in
the mid-1980s, in the setting of artificial intelligence, with the
goal of developing a simple yet rigorous approach to automated reasoning
from uncertain or prioritized incomplete information. Possibilistic
logic is especially adapted to automated reasoning when the
available information is pervaded by ambiguities. In fact,
possibilistic logic is a natural extension of classical logic in which
the notion of total order/partial order is embedded in the logic.

Possibilistic Logic is based on \emph{possibility theory}. Possibilistic theory, as its name implies, deals with the possible rather than probable values of a variable with possibility being a matter of degree. One merit of possibilistic theory is at one and the same time to represent imprecision (in the form of fuzzy sets) and quantity uncertainty (through the pair of numbers that measure \emph{possibility} and \emph{necessity}).

Our study in possibilistic logic is devoted to a fragment of possibilistic logic, in which knowledge bases are only \emph{necessity-quantified} statements. A  necessity-valued formula is a pair $(\varphi \; \alpha)$ in which $\varphi$ is a
classical logic formula and $\alpha \in (0,1]$ is a positive number.
The pair $(\varphi \; \alpha)$ expresses that the formula $\varphi$
is certain at least to the level $\alpha$, that is $N(\varphi) \geq
\alpha$, in which $N$ is a necessity measure modeling our possibly
incomplete state knowledge \cite{Dubois94}. $\alpha$ is not a
probability (like it is in probability theory), but it induces a
certainty (or confidence) scale. This value is determined by the
expert providing the knowledge base. A necessity-valued knowledge
base is then defined as a finite set (that is to say a conjunction) of
necessity-valued formul\ae.\

The following properties hold \wrt necessity-valued formul\ae:\

\begin{equation}\label{eq:poss-and}
    N(\varphi \wedge \psi ) = min(\{N(\varphi), N(\psi)\})
\end{equation}

\begin{equation}\label{eq:poss-or}
    N(\varphi \vee \psi ) \geq max(\{N(\varphi), N(\psi)\})
\end{equation}

\begin{equation}\label{eq:poss-inferences}
  \text{ if } \varphi \vdash \psi \text{ then } N(\psi)
  \geq N(\varphi)
\end{equation}

Dubois \etal, in \cite{Dubois94} introduced a formal system for
necessity-valued logic which is based on the following axioms
schemata (propositional case):

\begin{description}
\item[(A1)] $( \varphi  \rightarrow (\psi \rightarrow \varphi  ) \;  1)$

\item[(A2)] $((\varphi \rightarrow (\psi \rightarrow \xi )) \rightarrow (( \varphi \rightarrow \psi ) \rightarrow (\varphi \rightarrow \xi)) \; 1)$

\item[(A3)] $( (\neg \varphi \rightarrow \neg \psi ) \rightarrow ((\neg \varphi \rightarrow \psi) \rightarrow \varphi) \; 1 )$

\end{description}

\noindent Inference rules:

\begin{description}
\item[(GMP)] $(\varphi \; \alpha), (\varphi \rightarrow \psi \; \beta) \vdash (\psi \; min\{\alpha, \beta \} )$
\item[(S)] $(\varphi \; \alpha) \vdash (\varphi \; \beta)$ if $\beta \leq \alpha$
\end{description}

According to Dubois \etal, in \cite{Dubois94}, basically we need a
complete lattice to express the levels of uncertainty  in
Possibilistic Logic. Dubois \etal$~$ extended the axioms schemata and
the inference rules for considering partially ordered sets. We shall
denote by $\vdash_{PL}$ the inference under Possibilistic Logic
without paying attention to whether the necessity-valued formul\ae\ are
using  a totally ordered set or a partially ordered set for
expressing the levels of uncertainty.

The problem of inferring automatically the necessity-value of a
classical formula from a possibilistic base was solved by an
extended version of \emph{resolution} for possibilistic logic (see
\cite{Dubois94} for details).

One of the main principles of possibilistic logic is that:

\begin{remark}\label{remark:principlePossLogic}
The strength of a conclusion is the strength of the weakest argument
used in its proof.
\end{remark}

According to Dubois and Prade \cite{DuboisP04a}, the contribution of
possibilistic logic setting is to relate this principle (measuring
the validity of an inference chain by its weakest link) to fuzzy
set-based necessity measures in the framework of Zadeh's
possibilistic theory, since the following pattern then holds:

$$N(\sim p \vee q) \geq \alpha \textnormal{ and } N(p) \geq \beta \textnormal{ imply } N(q) \geq min(\alpha, \beta)$$

\noindent This interpretive setting provides a semantic
justification to the claim that the weight attached to a conclusion
should be the weakest among the weights attached to the formul\ae\
involved in the derivation.


\section{Syntax}\label{sec:Poss-syntax}

In this section, the general syntax for possibilistic disjunctive logic programs will be presented. This syntax is based on the
standard syntax of extended disjunctive logic programs (see Section
\ref{aspSyntaxis}).

We start by defining some concepts for managing the possibilistic values of a possibilistic knowledge base\footnote{Some concepts presented in this section extend
some terms presented in \cite{NicGarSteLef06}.}. We want to point out that in the whole document only finite lattices are considered.
This assumption was made based on the recognition that in real applications we will
rarely have an infinite set of labels for expressing the incomplete
state of a knowledge base.

A \emph{possibilistic atom} is a pair $p = (a, q) \in {\cal A} \times {\cal Q}$, in which ${\cal A}$ is a finite set of atoms and $({\cal Q},\leq)$ is a lattice. The projection $*$ to a possibilistic atom $p$ is defined as follows: $p^* = a$. Also given a set of
possibilistic atoms $S$, $*$ over $S$ is defined as follows: $S^* = \{ p^* | p \in S \}$.

Let $({\cal Q},\leq )$ be a lattice. A possibilistic disjunctive
clause $R$ is of the form:

$$\alpha : {\cal A} \leftarrow {\cal B}^+ , \ not \; {\cal B}^-$$

\noindent in which $\alpha \in {\cal Q}$ and ${\cal A} \leftarrow {\cal
B}^+ , \ not \; {\cal B}^-$ is an extended disjunctive clause as
defined in Section \ref{aspSyntaxis}. The projection $*$ for a possibilistic clause is $R^* = {\cal A} \leftarrow {\cal B}^+ , \ not \; {\cal B}^-$. On the other hand, the projection $n$ for a possibilistic clause is $n(R) = \alpha$. This projection denotes the degree of necessity captured by the certainty level of the information described by $R$. A possibilistic constraint $C$ is of the form:
$$\top_{\cal Q} : \; \; \;  \leftarrow {\cal B}^+ , \ not \; {\cal B}^-$$

\noindent in which $\top_{\cal Q}$ is the top of the lattice
$({\cal Q},\leq )$ and $ \leftarrow {\cal B}^+ , \ not \; {\cal
B}^-$ is a constraint as defined in Section \ref{aspSyntaxis}.
The projection $*$ for a possibilistic constraint $C$ is: $C^* = \;\; \leftarrow {\cal B}^+ , \ not \; {\cal
B}^-$. Observe that the possibilistic constraints have the top of the lattice $({\cal Q},\leq )$ as an uncertain value, this assumption is due to the fact that similar a constraint in standard ASP, the purpose of a
possibilistic constraint is to eliminate possibilistic models. Hence, it can be assumed that there is no doubt about the veracity of the information captured by a possibilistic constraint.
However, as in standard ASP, one can define possibilistic constraints of the form: $\alpha : x \leftarrow {\cal B}^+ , \ not \; {\cal B}^-, \; not \; x$ such that $x$ is an atom which is  not used in any other possibilistic clause and $\alpha \in {\cal Q}$. This means that the user can define possibilistic constraints with different levels of certainty.

A possibilistic disjunctive logic program $P$ is a tuple of the form
$\langle ({\cal Q},\leq), N \rangle$, in which $N$ is a finite set of
possibilistic disjunctive clauses and possibilistic constraints. The
generalization of $*$ over $P$ is as follows: $P^* = \{ r^* | r \in
N \}$. Notice that $P^*$ is an extended disjunctive program. When
$P^*$ is a normal program, $P$ is called a possibilistic normal
program. Also, when $P^*$ is a positive disjunctive program, $P$ is
called a possibilistic positive logic program and so on. A given set
of possibilistic disjunctive clauses $\{\gamma, \dots, \gamma \}$ is
also represented as $\{\gamma ; \dots ; \gamma \}$ to avoid
ambiguities with the use of the comma in the body of the clauses.

Given a possibilistic disjunctive logic program $P = \langle ({\cal
Q},\leq), N \rangle$, we define the \emph{$\alpha$-cut} and the
\emph{strict $\alpha$-cut} of $P$, denoted respectively by
$P_{\alpha}$ and
$P_{\overline{\alpha}}$, by\\

$P_{\alpha} = \langle ({\cal Q},\leq), N_{\alpha} \rangle$ such that
$N_{\alpha} = \{ c | c \in N \text{ and } n(c) \geq  \alpha \}$

$P_{\overline{\alpha}} = \langle ({\cal Q},\leq),
N_{\overline{\alpha}} \rangle$ such that $N_{\overline{\alpha}} = \{
c | c \in N \text{ and } n(c) >  \alpha \}$

\begin{example}\label{example:Assig-heart}
In order to illustrate a possibilistic program, let us go back to
our scenario described in Section \ref{sec:PossProg-Motivation}.
Let $({\cal Q}, \preceq)$ be the lattice of Figure
\ref{fig:lattice-background} such that the relation $A \preceq B$
means that $A$ is less possible than $B$. The possibilistic program
$P := \langle ({\cal Q},\preceq), N \rangle$ will be the following
set of possibilistic clauses:

\noindent It is probable that if the organ donor has an infection, then
the organ recipient can be infected or not after a graft:\\

\noindent  \textbf{probable}: $r\_inf(present, T2) \vee \lnot
r\_inf(present,T2)$ $\leftarrow$ $action(transplant, T),$ \\
$~~~~~~~~~~~~~~~~~~~~~~~~~~~~~~~~~~~~ d\_inf(present, T), T2 = T + 1.$ \\


\noindent 
It is confirmed that the organ's functions can be: good, delayed and terminal after a graft.\\

\noindent \textbf{confirmed}: $o(good\_graft\_funct, T2) \vee
o(delayed\_graft\_funct, T2)\vee$ \\
$o(terminal\_insufficient\_funct, T2) \leftarrow
 action(transplant, T),T2 = T + 1.$ \\

\noindent 
It is confirmed that if the organ's functions are terminally insufficient
then a transplanting is necessary. \\

\noindent \textbf{confirmed}: $action(transplant,T) \leftarrow$
$o(terminal\_insufficient\_funct, T)$.\\

\noindent 
It is plausible that the clinical situation of the organ recipient can be
stable if the functions of the graft are good. \\

\noindent \textbf{plausible}: $cs(stable, T) \leftarrow
o(good\_graft\_funct, T)$.\\

\noindent 
It is plausible that the clinical situation of the organ recipient can be
unstable if the functions of the graft are delayed. \\

\noindent \textbf{plausible}: $cs(unstable, T) \leftarrow
o(delayed\_graft\_funct, T)$.\\

\noindent 
It is plausible that the clinical situation of the organ recipient can be
of 0-urgency if the functions of the graft are terminally insufficient
after the graft. \\

\noindent \textbf{plausible}: $cs($0-urgency$, T2) \leftarrow
o(terminal\_insufficient\_funct, T2),$ \\
$~~~~~~~~~~~~~~~~~~~~~~~~~~~~~~~~~~~~ action(transplant,T), T2 = T + 1$.\\

\noindent 
It is certain that the doctor cannot do two
actions at the same time. \\

\noindent \textbf{certain}: $\;\; \leftarrow action(transplant, T),
action(wait, T)$. \\

\noindent 
It is certain that a transplant cannot be done if the organ recipient is dead. \\

\noindent \textbf{certain}: $\;\; \leftarrow action(transplant,T),
cs(dead,T)$. \\

\noindent The initial state of the automata of Figure
\ref{automata-actions-infections} is captured by the following
possibilistic clauses:\\

\noindent \textbf{certain}: $ d\_inf(present,0)$.

\noindent \textbf{certain}: $ \lnot r\_inf(present,0)$.

\noindent \textbf{certain}: $ o(terminal\_insufficient\_funct,0)$.

\noindent \textbf{certain}: $cs(stable, 0)$.

\end{example}

\section{Semantics}\label{sec:Poss-Sems}

In \S \ref{sec:Poss-syntax}, the syntax for any possibilistic disjunctive program was introduced, Now, in this section, a semantics for capturing these programs is studied. This semantics will be defined in terms of the standard definition of the answer set semantics (\S \ref{sec:ASP-semantics}) and the proof theory of possibilistic logic (\S \ref{subsect:PS}).

As sets of atoms are considered as interpretations, 
two basic operations between sets of possibilistic atoms are defined; also a relation of order between them is defined:
Given a finite set of atoms ${\cal A}$  and a lattice (${\cal Q}$,$\leq$), ${\cal PS'} = 2^{{\cal A} \times {\cal Q} }$ and
$${\cal PS} = {\cal PS'} \setminus \{A |  A \in {\cal PS} \text{ such that } x \in {\cal A} \text{ and } Cardinality(\{ (x, \alpha) | (x, \alpha) \in A \}) \geq 2 \}\footnote{$Cardinality$ is a function which returns the cardinality of a set.}$$

\noindent Observe that ${\cal PS'}$ is the finite set of all the possibilistic atom sets induced by ${\cal A}$ and $Q$. Informally speaking, ${\cal PS}$ is the subset of ${\cal PS'}$ such that each set of ${\cal PS}$ has no atoms with different uncertain value.

\begin{definition}
Let ${\cal A}$ be a finite set of atoms and (${\cal Q}$,$\leq$) be a
lattice.
$\forall A, B \in {\cal PS}$, we define.

\noindent\begin{tabular}{ll} $A \sqcap B$ & $= \{ (x, {\cal GLB}
(\{\alpha, \beta \}) | (x,\alpha) \in A \wedge (x, \beta) \in B \}$ \\

$A \sqcup B$ & $ = \{(x, \alpha) | (x, \alpha) \in A \; and \; x \notin B^* \}$
$\cup$\\
& $\{ (x, \alpha) | x \notin A^* \; and \; (x, \alpha) \in B \}$ $\cup$
\\ & $\{(x, {\cal LUB}(\{ \alpha , \beta \}) | (x, \alpha) \in A \text{ and } $  $(x, \beta) \in B \}$. \\

$A \sqsubseteq B$ & $\Longleftrightarrow A^* \subseteq B^* $, and $\forall x, \alpha, \beta, (x,\alpha) \in A \wedge  (x, \beta) \in B$ \\
& 
then $\alpha \leq \beta $.

\end{tabular}
\end{definition}

This definition is almost the same as Definition 7 presented in \cite{NicGarSteLef06}.
The main difference is that in Definition 7 from \cite{NicGarSteLef06} the operations $\sqcap$ and $\sqcup$ are defined in terms of the operators \emph{min} and \emph{max} instead of the operators ${\cal GLB}$ and ${\cal LUB}$. Hence, the following proposition is a direct result of Proposition 6 of \cite{NicGarSteLef06}.

\begin{proposition}
$({\cal PS},\sqsubseteq )$ is a complete lattice.
\end{proposition}

Before moving on, let us define the concept of \emph{i-greatest set} \wrt ${\cal PS}$ as follows: Given $M \in {\cal PS}$, $M$ is an \emph{i-greatest set} in ${\cal PS}$ iff $\nexists M' \in {\cal PS}$ such that  $M \sqsubseteq M'$. For instance, let ${\cal PS} = \{ \{ (a, 1)\}, \{(a, 2)\}, \{(a, 2), (b,1 )\} , \{ (a, 2), (b,2 ) \} \}$. One can see that ${\cal PS}$ has two i-greatest sets: $\{(a, 2)\}$ and  $\{ (a, 2), (b,2 ) \}$. The concept of i-greatest set will play a key role in the definition of possibilistic answer sets in order to infer possibilistic answer sets with optimal certainty values.

\subsection{Possibilistic answer set semantics}\label{sec:Poss-ASPSem}

Similar to the definition of answer set semantics, the possibilistic answer set semantics is defined in terms of a
syntactic reduction. This reduction is inspired by the Gelfond-Lifschitz reduction.

\begin{definition}[Reduction $P_M$]\label{def:reduction}
Let $P = \langle ({\cal Q},\leq), N \rangle$ be a possibilistic
disjunctive logic program, M be a set of atoms. $P$ reduced by $M$
is the positive possibilistic disjunctive logic program:\\

\noindent \begin{tabular}{ll} $P_M :=$ & $\{  (n(r) :{\cal A} \cap M
\leftarrow {\cal B}^+ ) | r \in N, {\cal A} \cap M \neq \emptyset,$
 $ {\cal B}^- \cap M = \emptyset, {\cal B}^+ \subseteq M \}$
\end{tabular}\\

\noindent in which $r^*$ is of the form $ {\cal A} \leftarrow {\cal
B}^+ , \ not \; {\cal B}^-$.
\end{definition}

Notice that $(P^*)_M$ is not exactly equal to the Gelfond-Lifschitz
reduction.
For instance, let us consider the following programs:\\

\begin{tabular}{llllll}
$P:$&& $P_{\{c,b\}}:$& & $(P^*)^{\{c,b\}}:$& \\
&$\alpha_1 : a \vee b$. & &$\alpha_1 : b$. & &$a \vee b$.\\
&$\alpha_2 : c \leftarrow \; not\; a$. & &$\alpha_2 : c$.  & & $ c$.\\
&$\alpha_3 : c \leftarrow \; not\; b$. & && \\
\end{tabular}\\

\noindent
The program $P_{\{c,b\}}$ is obtained from $P$ and $\{c,b\}$ by applying Definition \ref{def:reduction} and the program $(P^*)^{\{c,b\}}$ is obtained from  $P^*$ and  $\{c,b\}$  by applying the Gelfond-Lifschitz reduction. Observe that the reduction of Definition \ref{def:reduction} removes from the head of the possibilistic disjunctive clauses any atom which does not belong to $M$.
As we will see in Section \ref{sec:semPosFix-point}, this
property will be helpful for characterizing the possibilistic answer set in terms of a fixed-point operator. It is worth mentioning that the reduction $(P^*)_M$ also has a different effect from the Gelfond-Lifschitz reduction in the class of normal programs. This difference is illustrated in the following programs:\\

\begin{tabular}{llllll}
$P:$&& $P_{\{a\}}:$& & $(P^*)^{\{a\}}:$& \\
&$\alpha_1 : a \leftarrow \; not \; b$. & &$\alpha_1 : a$. & & $ a$.\\
&$\alpha_2 : a \leftarrow b$.          & &     & &$ a \leftarrow b$. \\
&$\alpha_3 : b \leftarrow c$.          & &     & &$  b \leftarrow c$.\\
\end{tabular}\\

\begin{example}\label{example:reduction}
Continuing with our medical scenario described in the
introduction, let $P$ be a ground instance of the possibilistic
program presented in Example \ref{example:Assig-heart}: \\

{\small \noindent  \textbf{probable}: $r\_inf(present, 1) \vee
no\_r\_inf(present,1)$ $\leftarrow$ $action(transplant, 0),$ \\
$~~~~~~~~~~~~~~~~~~~~~~~~~~~~~~~~~~~~~~~~~~~~~~~ d\_inf(present, 0).$ \\
\noindent \textbf{confirmed}: $o(good\_graft\_funct, 1) \vee
o(delayed\_graft\_funct, 1)\vee$ \\
$~~~~~~~~~~~~~~~~~~~~~~ o(terminal\_insufficient\_funct, 1)
\leftarrow  action(transplant, 0).$ \\
\noindent \textbf{confirmed}: $action(transplant,0) \leftarrow$
$o(terminal\_insufficient\_funct, 0)$.\\
\noindent \textbf{plausible}: $cs(stable, 1) \leftarrow
o(good\_graft\_funct, 1)$.\\
\noindent \textbf{plausible}: $cs(unstable, 1) \leftarrow
o(delayed\_graft\_funct, 1)$.\\
\noindent \textbf{plausible}: $cs($0-urgency$, 1) \leftarrow
o(terminal\_insufficient\_funct, 1),$ \\
$~~~~~~~~~~~~~~~~~~~~~~~~~~~~~~~~~~~~~~~~~~~~~~ action(transplant,0)$.\\
\noindent \textbf{certain}: $\;\; \leftarrow action(transplant, 0),
action(wait, 0)$. \\
\noindent \textbf{certain}: $\;\; \leftarrow action(transplant,0),
cs(dead,0)$. \\
\noindent \textbf{certain}: $ d\_inf(present,0)$. \\
\noindent \textbf{certain}: $ no\_r\_inf(present,0)$. \\
\noindent \textbf{certain}: $ o(terminal\_insufficient\_funct,0)$. \\
\noindent \textbf{certain}: $cs(stable, 0)$. \\
}

\noindent Observe that the variables of time $T$ and $T2$ were
instantiated with the values $0$ and $1$ respectively; moreover,
observe that
the atoms $\lnot r\_inf(present,0)$ and $\lnot r\_inf(present,1)$
were replaced by $no\_r\_inf(present,0)$ and $no\_r\_inf(present,1)$
respectively. This change was applied in order to manage the strong
negation, $\lnot$ .

Now, let $S$ be the following possibilistic set: \\

\noindent
\begin{tabular}{ll}
$S =$ & $\{(d\_inf(present,0), certain),$ $(no\_r\_inf(present,0), certain),$ \\
& $(o(terminal\_insufficient\_funct,0),certain),$  $(cs(stable,0), certain),$ \\
& $(action(transplant,0), confirmed),$ $(o(good\_graft\_funct,1), confirmed),$ \\
& $(cs(stable,1), plausible ),$ $(no\_r\_inf(present,1), probable)\}$. \\
\end{tabular} \\

\noindent One can see that $P_{S^*}$ is:\\

{\small \noindent  \textbf{probable}:
$no\_r\_inf(present,1)$ $\leftarrow$ $action(transplant, 0), d\_inf(present, 0).$ \\
\noindent \textbf{confirmed}: $o(good\_graft\_funct, 1)
\leftarrow  action(transplant, 0).$ \\
\noindent \textbf{confirmed}: $action(transplant,0) \leftarrow$
$o(terminal\_insufficient\_funct, 0)$.\\
\noindent \textbf{plausible}: $cs(stable, 1) \leftarrow
o(good\_graft\_funct, 1)$.\\
\noindent \textbf{plausible}: $cs(unstable, 1) \leftarrow
o(delayed\_graft\_funct, 1)$.\\
\noindent \textbf{plausible}: $cs($0-urgency$, 1) \leftarrow
o(terminal\_insufficient\_funct, 1),$ \\
$~~~~~~~~~~~~~~~~~~~~~~~~~~~~~~~~~~~~~~~~~~~~~~ action(transplant,0)$.\\
\noindent \textbf{certain}: $\;\; \leftarrow action(transplant, 0),
action(wait, 0)$. \\
\noindent \textbf{certain}: $\;\; \leftarrow action(transplant,0),
cs(dead,0)$. \\
\noindent \textbf{certain}: $ d\_inf(present,0)$. \\
\noindent \textbf{certain}: $ no\_r\_inf(present,0)$. \\
\noindent \textbf{certain}: $ o(terminal\_insufficient\_funct,0)$. \\
\noindent \textbf{certain}: $cs(stable, 0)$. \\
}
\end{example}

Once a possibilistic logic program $P$ has been reduced by a set of possibilistic atoms $M$, it is possible to test whether $M$ is a possibilistic answer set of the program $P$. For this end, we consider a syntactic approach; meaning that it is based on the proof theory of possibilistic logic. Let us remember that the possibilistic logic is axiomatizable \cite{Dubois94}; hence, the inference in possibilistic logic can be managed by both a syntactic approach  (axioms and inference rules) and a possibilistic model theory approach (interpretations and possibilistic distributions).


Since the certainty value of a possibilistic disjunctive clause can belong to a partially ordered set, the inference rules of possibilistic logic introduced in Section \ref{subsect:PS} have to be generalized in terms of bounds. The generalization of \textbf{GMP} and \textbf{S} is defined as follows:

\begin{description}
\item[(GMP*)] $(\varphi \; \alpha), (\varphi \rightarrow \psi \; \beta) \vdash (\psi \; GLB\{\alpha, \beta \} )$
\item[(S*)] $(\varphi \; \alpha), (\varphi \; \beta) \vdash (\varphi \; \gamma)$, where  $\gamma  \leq GLB\{\alpha,\beta\}$
\end{description}

\noindent Observe that these inference rules are essentially the same as the inference rules introduced in Section  \ref{subsect:PS}; however, they are defined in terms of $GLB$ to lead with certainty values which are not comparable (in Example \ref{example:PossAnswerSet} these inference rules are illustrated).


Once we have defined $GMP^*$ and $S^*$,  the inference $\Vvdash_{PL}$ is defined as follows:

\begin{definition}
Let $P =  \langle ({\cal Q},\leq), N \rangle$ be a possibilistic disjunctive logic program and $M \in {\cal PS}$.
\begin{itemize}
  \item We write $P \Vvdash_{PL} M$ when $M^*$ is an answer set of $P^*$ and $P_{M^*} \vdash_{PL} M$.
\end{itemize}

\end{definition}

One can see that $\Vvdash_{PL}$ is defining a joint inference between \emph{the answer set semantics} and \emph{the proof theory of possibilistic logic}.
Let us consider the following example.
\begin{example}\label{example:ASP-LP-inference}
Let $P =  \langle ({\cal Q},\leq), N \rangle$ be a possibilistic disjunctive logic program such that ${\cal Q} = \{0.1, \dots,$ $0.9\}$, $\leq$ denotes the standard relation in real numbers and $N$ is the following set of possibilistic clauses:

\begin{tabular}{lllll}
$0.6: a \vee b.$ & $~~~~~~~~$& $0.4: a \leftarrow not \; b. $ & $~~~~~~~~$& $0.8: b \leftarrow not \; a.$
\end{tabular}

\noindent It is easy to see that $P^*$ has two answer sets: $\{a\}$ and $\{b\}$. On the other hand, one can see that $P_{\{a\}} \vdash_{PL} \{(a, 0.6)\}$, $P_{\{a\}} \vdash_{PL} \{(a, 0.4)\}$, $P_{\{b\}} \vdash_{PL} \{(b, 0.6)\}$ and $P_{\{b\}} \vdash_{PL} \{(b, 0.8)\}$. This means that $P \Vvdash_{PL} \{(a, 0.6)\}$, $P \Vvdash_{PL} \{(a, 0.4)\}$, $P \Vvdash_{PL} \{(b, 0.6)\}$ and $P \Vvdash_{PL} \{(b, 0.8)\}$.
\end{example}

The basic idea of $\Vvdash_{PL}$ is to identify candidate sets of possibilistic atoms in order to consider them as possibilistic answer sets. The following proposition formalizes an important property of $\Vvdash_{PL}$.

\begin{proposition}\label{prop:ASPPossInference-LUB}
Let $P =  \langle ({\cal Q},\leq), N \rangle$ be a possibilistic disjunctive logic program and $M_1,M_2 \in {\cal PS}$ such that $M_1^* = M_2^*$. If $P \Vvdash_{PL} M1$ and $P \Vvdash_{PL} M2$, then $P \Vvdash_{PL} M_1 \sqcup M_2$.
\end{proposition}

In this proposition, since $M_1$ and $M_2$ are two sets of possibilistic atoms, ${\cal LUB}$ is instantiated in terms of $\sqsubseteq$. By considering  $\Vvdash_{PL}$ and the concept of \emph{i-greatest set}, a possibilistic answer set is defined as follows:

\begin{definition}[A possibilistic answer set]
Let $P =  \langle ({\cal Q},\leq), N \rangle$ be a possibilistic disjunctive logic program and $M$ be a set of possibilistic atoms
such that $M^*$ is an answer set of $P^*$. $M$ is a possibilistic answer set of $P$ iff $M$ is an i-greatest set in ${\cal PS}$ such that $P \Vvdash_{PL} M$.
\end{definition}

Essentially, a possibilistic answer set is an i-greatest set which is inferred by  $\Vvdash_{PL}$. In other words, a possibilistic answer set  is an answer set with \emph{optimal certainty values}. For instance, in Example \ref{example:ASP-LP-inference}, we saw that $P \Vvdash_{PL} \{(a, 0.6)\}$, $P \Vvdash_{PL} \{(a, 0.4)\}$, $P \Vvdash_{PL} \{(b, 0.6)\}$ and $P \Vvdash_{PL} \{(b, 0.8)\}$; however,  $\{(a, 0.4)\}$ and $\{(b, 0.6)\}$ are not i-greatest sets. This means that the possibilistic answer sets of the possibilistic program $P$ of Example \ref{example:ASP-LP-inference} are:   $\{(a, 0.6)\}$  and $\{(b, 0.8)\}$.

\begin{example}\label{example:PossAnswerSet}
Let $P$ be again the possibilistic program of Example
\ref{example:Assig-heart} and $S$ be the possibilistic set of atoms
introduced in Example \ref{example:reduction}.

One can see that $S^*$ is an answer set of the extended disjunctive program
$P^*$. Hence, in order to prove that $P \Vvdash_{PL} S$, we have to verify that $P_{S^*} \vdash_{PL} S$. This
means that for each possibilistic atom $p \in S$, $P_{S^*}
\vdash_{PL} p$. It is clear that \\

\begin{tabular}{ll}
$P_{S^*} \vdash_{PL}$ & $\{ (d\_inf(present,0), certain),$
$(no\_r\_inf(present,0), certain),$ \\
& $(o(terminal\_insufficient\_funct,0),certain),$ \\
&$(cs(stable,0), certain) \}$
\end{tabular}\\

\noindent Now let us prove $(cs(stable,1), plausible )$ from $P_{S^*}$. \\

{\small

\noindent
\begin{tabular}{lr}
 \textbf{Premises from $P_{S^*}$} & \\
1. $o(terminal\_insufficient\_funct,0)$ &  $certain$   \\
2. $o(terminal\_insufficient\_funct, 0) \rightarrow action(transplant,0)$ &  $confirmed$   \\
3. $action(transplant, 0) \rightarrow o(good\_graft\_funct, 1)$ & $confirmed$   \\
4. $o(good\_graft\_funct,1) \rightarrow cs(stable, 1)$ & $plausible$   \\
\textbf{From 1 and 2 by GMP*} & \\
5. $action(transplant,0)$ &  $confirmed$ \\
\textbf{From 3 and 5 by GMP*} & \\
6. $o(good\_graft\_funct, 1)$ & $confirmed$ \\
\textbf{From 4 and 6 by GMP*} & \\
7. $cs(stable, 1)$. & $plausible$ \\
\end{tabular}}\\

\noindent In this proof, we can also see the inference of the
possibilistic atom $(action(transplant,0),$ $confirmed)$. The proof
of the possibilistic atom $(no\_r\_inf(present,1), probable)$ is
similar to the proof of the possibilistic atom $(cs(stable,1),
plausible )$. Therefore, $P_{S^*} \vdash_{PL} S$ is
true. Notice that a possibilistic set
$S'$ such that $S'\neq S$, $P_{(S')^*} \vdash_{PL} S'$ and $S
\sqsubseteq S'$ does not exists; hence, $S$ is an i-greatest set. Then, $S$ is a possibilistic answer set of $P$.

By considering the possibilistic answer set $S$, what can we conclude about our medical scenario from $S$? We can conclude that if it is \emph{confirmed} that a transplant is performed on a
donor with an infection, it is \emph{probable} that the recipient
will not be infected after the transplant; moreover it is
\emph{plausible} that he will be stable. It is worth mentioning that
this \emph{optimistic} conclusion is just one of the possible
scenarios that we can infer from the program $P$. In fact, the
program $P$ has six possibilistic answer sets in which we can find
pessimistic scenarios such as it is \emph{probable} that the
recipient will be infected by the organ donor's infection and; moreover,
it is \emph{confirmed} that the recipient needs another
transplant.
\end{example}

Now, let us identify some properties of the possibilistic answer set
semantics. First, observe that there is an important condition \wrt
the definition of a \emph{possibilistic answer set} which is introduced by $\Vvdash_{PL}$: a possibilistic set $S$ cannot be a possibilistic answer set of a possibilistic logic program $P$ if $S^*$ is not an answer set of the
extended logic program $P^*$. This condition guarantees that any
clause of $P^*$ is satisfied by $S^*$. For instance, let us consider
the possibilistic logic program $P$:

\begin{tabular}{lll}\\
$0.4: a. $ &$~~~~~~~$& $0.6: b.$ \\
\end{tabular}\\

\noindent and the possibilistic set $S = \{ (a, 0.4)\}$. We can see
that $P_{S^*} \vdash_{PL} S$; however, $S^*$ is not an answer set of
$P^*$. Therefore, $P \Vvdash_{PL} S$ is \emph{false}. Then $S$ could not be a possibilistic answer set of $P$. This suggests, a direct relationship between the possibilistic answer semantics and the answer set semantics.

\begin{proposition}\label{prop:PostASP-ASP}
Let $P$ be a possibilistic disjunctive logic program. If $M$ is a
possibilistic answer set of P then $M^*$ is an answer set of $P^*$.
\end{proposition}

When all the possibilistic clauses of a possibilistic program $P$
have the same certainly level, the answer sets of $P^*$ can be directly generalized to the possibilistic answer sets of $P$.

\begin{proposition}\label{prop:Post-top-ASP-ASP}
Let $P = \langle ({\cal Q},\leq), N \rangle$ be a possibilistic disjunctive logic program and
$\alpha$ be a fixed element of ${\cal Q}$.
If $\forall r \in P$, $n(r) = \alpha$  and $M'$ is an answer set of $P^*$, then $M :=\{
(a, \alpha)| a \in M' \}$ is a possibilistic answer set of $P$.
\end{proposition}

For the class of possibilistic normal logic programs which are defined with a totally ordered set, our definition of possibilistic answer
set is closely related to the definition of a \emph{possibilistic
stable model} presented in \cite{NicGarSteLef06}. In fact, both
semantics coincide.

\begin{proposition}\label{prop:PosstASP-PossStable}
Let $P := \langle (Q,\leq), N \rangle$ be a possibilistic normal
program such that $(Q,\leq)$ is a totally ordered set and ${\cal
L}_{P}$ has no extended atoms. $M$ is a possibilistic answer set of
P if and only if $M$ is a possibilistic stable model of $P$.
\end{proposition}

To prove that the possibilistic answer set semantics is
computable, we will present an algorithm for computing possibilistic answer sets. With this in mind, let us remember that a classical resolvent is defined as follows: Assume that $C$ and $D$ are two clauses in their disjunctive form such that $C = a \vee l_1 \vee \dots \vee l_n$ and $D = \sim a \vee ll_1 \vee \dots \vee ll_m$. The clause $l_1 \vee \dots \vee l_n \vee  ll_1 \vee \dots \vee ll_m$ is called a resolvent of $C$ and $D$ \wrt $a$. Thus clauses $C$ and $D$ have a resolvent in case a literal $a$ exists such that $a$ appears in $C$ and $\sim a$ appears in $D$ (or conversely).

Now, let us consider a straightforward generalization of the possibilistic resolution rule introduced in \cite{Dubois94}:

\begin{description}
\item[(R)]   $(c_1 \; \alpha_1) (c_2 \; \alpha_2) \vdash (R(c_1,c_2) \; {\cal GLB}(\{\alpha_1, \alpha_2\}))$
\end{description}

\noindent in which $R(c_1,c_2)$ is any classical resolvent of $c_1$ and
$c_2$ such that $c_1$ and $c_2$ are disjunctions of literals. It is
worth mentioning that it is easy to transform any possibilistic
disjunctive logic program $P$ into a set of possibilistic
disjunctions ${\cal C}$. Indeed, ${\cal C}$ can be
obtained as follows:\\

\noindent ${\cal C} := \bigcup \{ (a_1 \vee \ldots \vee a_m \vee
\sim a_{m + 1} \vee \dots \vee \sim a_{j} \vee a_{j+1} \vee \dots, a_n \;
\alpha)| $ \\ $~~~~~~~~~~~~~~~~~~~~~ (\alpha: a_1 \vee \ldots \vee
a_m \leftarrow a_{m + 1}, \dots, a_{j}, not \; a_{j+1}, \dots, not\; a_n )
\in P \}$\\

\noindent Let us remember that whenever a possibilistic program is considered as a possibilistic theory, each negative literal $not \; a$
is replaced by $\sim a$ such that $\sim$ is regarded as the negation
in classic logic --- in Example \ref{example:poss-resolution}, the
transformation of a possibilistic program into a set of
possibilistic disjunctions is shown.

The following proposition shows that the resolution rule (R) is
sound.

\begin{proposition}\label{prop:poss-R-sound}
Let ${\cal C}$ be a set of possibilistic disjunctions, and $C = (c
\; \alpha)$ be a possibilistic clause obtained by a finite number of
successive application of \emph{(R)} to ${\cal C}$; then ${\cal C}
\vdash_{PL} C$.
\end{proposition}



Like the possibilistic rule introduced in \cite{Dubois94}, (R) is
complete for refutation. We will say that a possibilistic
disjunctive program $P$ is \emph{consistent} if $P$ has at least a
possibilistic answer set. Otherwise $P$ is said to be
\emph{inconsistent}. The degree of inconsistency  of a possibilistic
logic program $P$ is $Inc(P) = {\cal GLB}(\{ \alpha | P_{\alpha}
\text{ is consistent }\})$.

\begin{proposition}\label{prop:poss-R-Complete}
Let $P$ be a set of possibilistic clauses and ${\cal C}$ be the set
of possibilistic disjunctions obtained from $P$; then the valuation
of the optimal refutation by resolution from ${\cal C}$  is the
inconsistent degree of $P$.
\end{proposition}

The main implication of Proposition \ref{prop:poss-R-sound} and
Proposition \ref{prop:poss-R-Complete} is that (R) suggests a method
for inferring a possibilistic formula from a possibilistic knowledge
base.

\begin{corollary}\label{corrollary:optimal-PossValue}
Let $P := \langle ({\cal Q},\leq), N \rangle$ be a possibilistic
disjunctive logic program, $\varphi$ be a literal and ${\cal C}$ be a
set of possibilistic disjunctions obtained from $N \cup \{ ( \sim
\varphi \; \top_{\cal Q} ) \}$; then the valuation of the
optimal refutation from ${\cal C}$ is $n(\varphi)$, that is \; $P
\vdash_{PL} (\varphi \; n(\varphi)) $.
\end{corollary}



Based on the fact that the resolution rule (R) suggests a method for
inferring the necessity value of a possibilistic formula, we can
define the following function for computing the possibilistic answer
sets of a possibilistic program $P$. In this function, $\square$ denotes an empty clause.
\\

\noindent \textbf{Function} $Poss\_Answer\_Sets(P)$

\noindent Let $ASP(P^*)$ be a function that computes the answer set
models of the standard logic program $P^*$, for example DLV \cite{DLV}.

Poss-ASP $:= \emptyset$

For all $S \in ASP(P^*)$

\indent \indent Let ${\cal C}$ be the set of possibilistic
disjunctions obtained from $P_S$. \\
\indent \indent $S' := \emptyset$ \\
\indent \indent for all $a \in S$ \\
\indent \indent \indent $C' := {\cal C} \cup \{ ( \sim a \; \top_{\cal Q} ) \}$ \\
\indent \indent \indent Search for a deduction of $( R(\square) \; \alpha)$ by applying repeatedly \\
\indent \indent \indent the resolution rule (R) from $C'$, with
$\alpha$ maximal. \\
\indent \indent \indent $S' := S' \cup \{(a \; \alpha) \}$ \\
\indent \indent endfor \\
\indent \indent Poss-ASP $:= $ Poss-ASP $\cup \; S'$ \\
\indent endfor \\
\noindent \textbf{return}(Poss-ASP). \\

The following proposition proves that the function
$Poss\_Answer\_Sets$ computes all the possibilistic answer sets of a
possibilistic logic program.

\begin{proposition}\label{prop:algorithm-PossASP}
Let $P := \langle ({\cal Q},\leq), N \rangle$ be a possibilistic
logic program. The set Poss-ASP returned by $Poss\_Answer\_Sets(P)$
is the set of all the possibilistic answer sets of $P$.
\end{proposition}

In order to illustrate this algorithm,
let us consider the following example:

\begin{example}\label{example:poss-resolution}
Let $P := \langle ({\cal Q},\leq), N \rangle$ be a possibilistic
program such that ${\cal Q} := \{0,$ $0.1$, $\dots$, $0.9,$ $1 \}$, $\leq$ is the standard
relation between rational numbers and $N$ the following set of
possibilistic clauses:

\begin{tabular}{rrl}\\
$0.7: $ & $a \vee b $ & $ \leftarrow not \; c.$ \\
$0.6: $ & $c $ & $ \leftarrow not \; a, not \; b.$ \\
$0.8: $ & $a $ & $ \leftarrow b.$ \\
$0.9: $ & $e $ & $ \leftarrow b.$ \\
$0.6: $ & $b $ & $ \leftarrow a.$ \\
$0.5: $ & $b $ & $ \leftarrow a.$
\end{tabular}\\

\noindent First of all, we can see that $P^*$ has two answer sets:
$S_1: = \{a, b, e\}$ and $S_2:= \{c\}$. This means that $P$ has two
possibilistic answer set models. Let us consider $S_1$ for our
example. Then, one can see that $P_{S_1}$ is:

\begin{tabular}{rrl}\\
$0.7: $ & $ a \vee b$. &  \\
$0.8: $ & $a $ & $ \leftarrow b.$ \\
$0.9: $ & $e $ & $ \leftarrow b.$ \\
$0.6: $ & $b $ & $ \leftarrow a.$ \\
$0.5: $ & $b $ & $ \leftarrow a.$
\end{tabular}\\

\noindent Then ${\cal C} := \{ (a \vee b \; 0.7), (a \vee \sim b \;
0.8), (e \vee \sim b \; 0.9), (b \vee \sim a \; 0.6), (b \vee \sim a
\; 0.5)\}$. In order to infer the necessity value of the atom $a$,
we add $(\sim a \; 1)$ to ${\cal C}$ and a search for finding an
optimal refutation is applied. As we can see in Figure
\ref{fig:search-refutation-a}, there are three refutations, however
the optimal refutation is $(\square \; 0.7)$. This means that the
best necessity value for the atom $a$ is $0.7$.

\begin{figure}[th]
\begin{center}
\epsfig{file=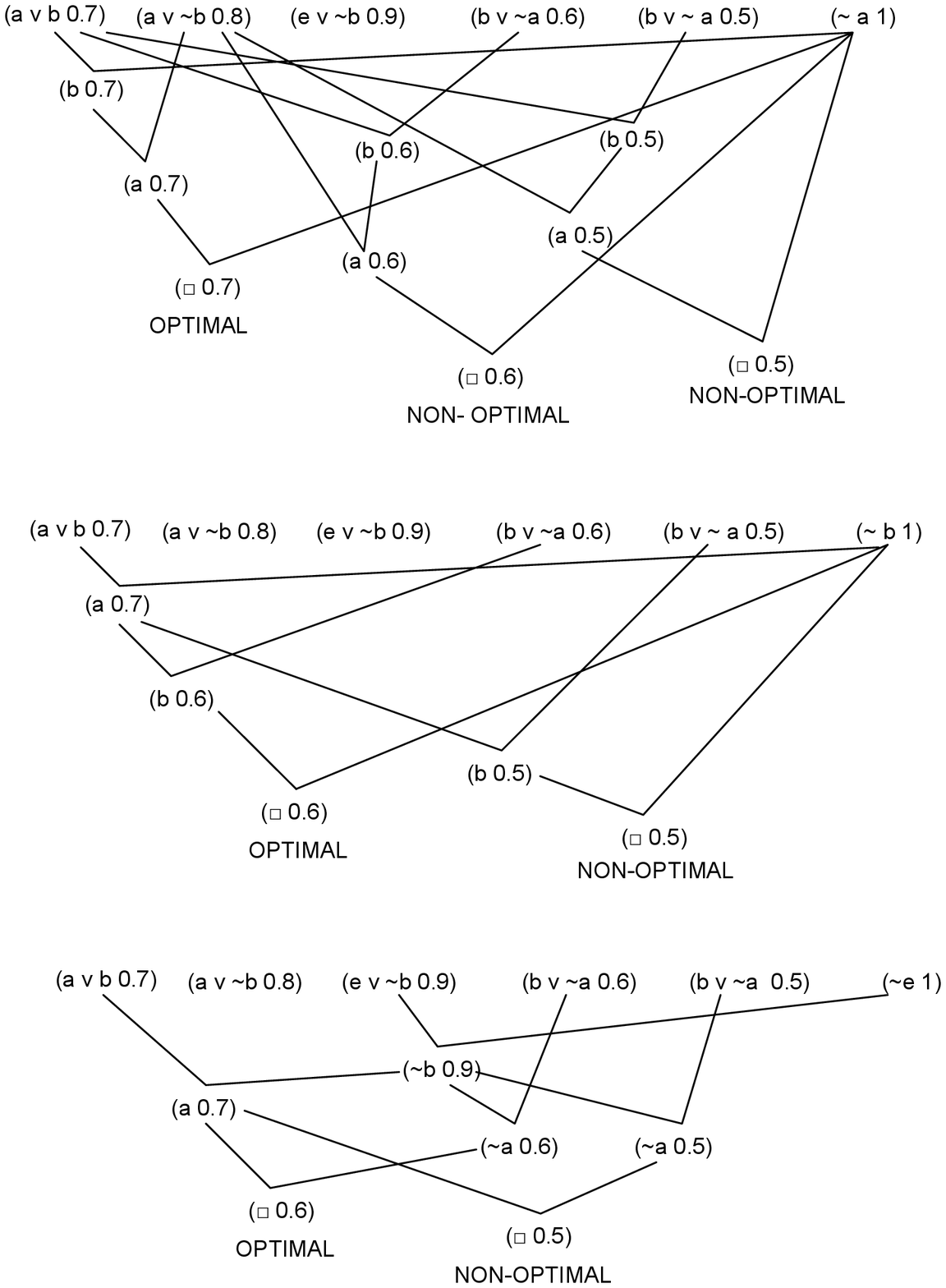,scale=0.6, bbllx=72, bblly=545,
bburx=474, bbury=754, clip=} \caption{Possibilistic resolution:
Search for an \emph{optimal refutation} for the atom $a$.
}\label{fig:search-refutation-a}
\end{center}
\end{figure}


In Figure \ref{fig:search-refutation-b}, we can see the optimal
refutation search for the atom $b$. As we can see the optimal
refutation is $(\square \; 0.6)$; hence the best necessity value for
the atom $b$ is $0.6$.

\begin{figure}[th]
\begin{center}
\epsfig{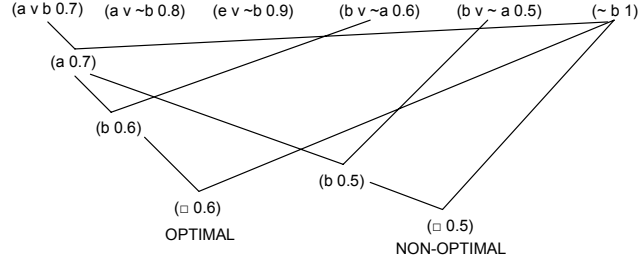} \caption{Possibilistic resolution:
Search for an \emph{optimal refutation} for the atom $b$.}
    \label{fig:search-refutation-b}
\end{center}
\end{figure}


In Figure \ref{fig:search-refutation-e}, we can see that the best necessity value for the atom $e$ is $0.6$.


\begin{figure}[th]
\begin{center}
\epsfig{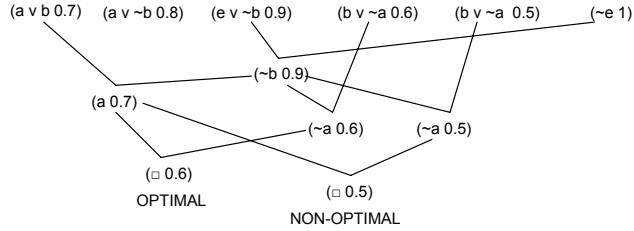} \caption{Possibilistic resolution:
Search for an \emph{optimal refutation} for the atom $e$. }
    \label{fig:search-refutation-e}
\end{center}
\end{figure}

Thought the search, we can infer that a possibilistic answer
set of the program $P$ is : $\{(a, 0.7), (b,0.6), (e, 0.6)\}$.

\end{example}

\subsection{Possibilistic answer sets based on partial
evaluation}\label{sec:semPosFix-point}

We have defined a possibilistic answer set semantics by considering
the formal proof theory of possibilistic logic. However, in standard
logic programming there are several frameworks for analyzing,
defining and computing logic programming semantics
\cite{Dix95,Dix95a}. One of these approaches is based on program
transformations, in fact there are many studies on this approach, for example \cite{bradix99,BraDix97,BraDix98,DixOsoZep01}. For the case of
disjunctive logic program, one important transformation is
\emph{partial evaluation (also called unfolding)} \cite{bradix99}.

This section shows that it is also possible to define a
possibilistic disjunctive semantics based on an operator which is a
combination between partial evaluation for disjunctive logic
programs and the infer rule $GMP^*$ of possibilistic logic. This semantics has the same behavior as the
semantics based on the proof theory of possibilistic logic.

This section starts by defining a version of the general
principle of partial evaluation (GPPE) for possibilistic positive
disjunctive clauses.

\begin{definition}[Grade-GPPE (G-GPPE)]
Let $r_1$ be a possibilistic clause of the form $\alpha : {\cal A}
\leftarrow {\cal B}^+ \cup \{ B \}$ and $r_2$ a possibilistic clause
of the form $\alpha_1 : {\cal A}_1$  such that  $B \in {\cal A}_1$ and $B \notin B^+$, then \\

$\text{G-GPPE}(r_1,r_2)  = ({\cal GLB}( \{ \alpha , \alpha_1 \}):
{\cal A} \cup ({\cal A}_1 \setminus \{B \} ) \leftarrow {\cal B}^+
)$
\end{definition}

Observe that one of the possibilistic clauses which is considered by
G-GPPE has an empty body. For instance, let us consider the
following two possibilistic clauses: \\

$r_1 = \; 0.7 : a \vee b$.\\
\indent $r_2 = \;0.9 : e \leftarrow b$. \\

\noindent Then G-GPPE$(r_1,r_2) = (0.7 : e \vee a)$.
Now, by considering G-GPPE, we will define the operator ${\cal T}$.

\begin{definition} Let $P$ be a possibilistic positive logic
program. The operator ${\cal T}$ is defined as follows:

\[
{\cal T}(P) := P \cup \{ \text{G-GPPE}(r_1,r_2) | r_1, r_2 \in P\}
\]


\end{definition}

In order to illustrate the operator ${\cal T}$, let us consider the
program $P_{S_1}$ of Example \ref{example:poss-resolution}.

\begin{tabular}{rrl}\\
$0.7: $ & $ a \vee b$. &  \\
$0.8: $ & $a $ & $ \leftarrow b.$ \\
$0.9: $ & $e $ & $ \leftarrow b.$ \\
$0.6: $ & $b $ & $ \leftarrow a.$ \\
$0.5: $ & $b $ & $ \leftarrow a.$
\end{tabular}\\

\noindent Hence, ${\cal T}(P_{S_1})$ is:

\begin{tabular}{rlcrl}\\
$0.7: $ & $ a \vee b$. & $~~~$ &  $0.7: $ & $a$. \\
$0.8: $ & $a  \leftarrow b.$ & & $0.7: $  & $ e \vee a$. \\
$0.9: $ & $e  \leftarrow b.$ & &$0.6: $  & $b$. \\
$0.6: $ & $b \leftarrow a.$ & &$0.5: $  &$b$. \\
$0.5: $ & $b \leftarrow a.$ & & &  \\
\end{tabular}\\

\noindent Notice that by considering the possibilistic clauses that
were added to $P^{S_1}$ by ${\cal T}$, one can reapply G-GPPE.
For instance, if we consider $0.6: b$ and $0.9: e
\leftarrow b$ from ${\cal T}(P^{S_1})$, G-GPPE infers $0.6: e$. Indeed, ${\cal T}({\cal T}(P_{S_1}))$ is:

\begin{tabular}{rlcrlcrl}\\
$0.7: $ & $ a \vee b$. & $~~~$ &  $0.7: $ & $a$. & $~~~$ & $0.6: $ & $a$. \\
$0.8: $ & $a  \leftarrow b.$ & & $0.7: $  & $ e \vee a$. & &
$0.5: $ & $a$. \\
$0.9: $ & $e  \leftarrow b.$ & &$0.6: $  & $b$. & &
$0.6: $ & $e$. \\
$0.6: $ & $b \leftarrow a.$ & &$0.5: $  &$b$. & &
$0.5: $ & $e$. \\
$0.5: $ & $b \leftarrow a.$ & & &  & &
$0.6: $ & $b \vee e$. \\
 &  & & &  & & $0.5: $ & $b \vee e$. \\
\end{tabular}\\

An important property of the operator ${\cal T}$ is that it always
reaches a fixed-point.

\begin{proposition}\label{prop:FixPointOperatorPossSem}
Let $P$ be a possibilistic disjunctive logic program. If $\Gamma_0
:= {\cal T}(P)$ and $\Gamma_i := {\cal T}(\Gamma_{i-1})$ such that
$i \in {\cal N}$, then $ \exists \; n \in {\cal N}$ such that
$\Gamma_n = \Gamma_{n-1}$. We denote $\Gamma_n$ by $\Pi(P)$.
\end{proposition}

Let us consider again the possibilistic program $P_{S_1}$. We can
see that $\Pi(P_{S_1})$ is:

\noindent
\begin{tabular}{rlcrlcrlcrl}\\
$0.7: $ & $ a \vee b$. & &  $0.7: $ & $a$. &  & $0.6: $ & $a$. &  & $0.6$ & $a \vee e.$ \\
$0.8: $ & $a  \leftarrow b.$ & & $0.7: $  & $ e \vee a$. & & $0.5: $ & $a$. & & $0.5$ & $a \vee e.$ \\
$0.9: $ & $e  \leftarrow b.$ & &$0.6: $  & $b$. & &
$0.6: $ & $e$. & & &\\
$0.6: $ & $b \leftarrow a.$ & &$0.5: $  &$b$. & &
$0.5: $ & $e$. & & & \\
$0.5: $ & $b \leftarrow a.$ & & &  & &
$0.6: $ & $b \vee e$.  & & &\\
 &  & & &  & & $0.5: $ & $b \vee e$. & & & \\
\end{tabular}\\

Observe that in $\Pi(P_{S_1})$ there are possibilistic facts
(possibilistic clauses with empty bodies and one atom in their
heads) with different necessity value. In order to infer the optimal
necessity value of each possibilistic fact, one can consider the
\emph{least upper bound} of these values. For instance, the optimal
necessity value for the possibilistic atom $a$ is ${\cal LUB}(\{0.7,
0.6,0.5\}) = 0.7$. Based on this idea, $Sem_{min}$ is defined as follows.

\begin{definition}
Let $P$ be a possibilistic logic program and $Facts(P,a) := \{
(\alpha : a) | (\alpha : a) \in P
\}$. $Sem_{min}(P) := \{ (x,\alpha) | Facts(P,x) \neq \emptyset
\text{ and }$ $ \alpha := {\cal LUB}( \{ n(r) | {r\in Facts(P,x)}
\}) \}$ in which $x \in {\cal L}_P$.
\end{definition}

It is easy to see that $Sem_{min}(\Pi (P_{S_1}))$ is $\{ (a, 0.7),
(b, 0.6), (e,0.6)\}$. Now by considering the operator ${\cal T}$ and
$Sem_{min}$, we can define a semantics for possibilistic disjunctive
logic programs that will be called possibilistic-${\cal T}$ answer
set semantics.

\begin{definition}\label{def:SemPossFixPoint}
Let $P$ be a possibilistic disjunctive logic program and $M$ be a
set of possibilistic atoms such that $M^*$ is an answer set of
$P^*$. $M$ is a possibilistic-${\cal T}$ answer set of P if and only
if  $M = Sem_{min}( \Pi (P_{M^*}))$.
\end{definition}

In order to illustrate this definition, let us consider again the
program $P$ of Example \ref{example:poss-resolution} and $S = \{ (a,
0.7), (b, 0.6), (e, 0.6)\}$. As commented in Example
\ref{example:poss-resolution},  $S^*$ is an answer set of $P^*$. We
have already seen that $Sem_{min}(\Pi (P_{S_1}))$ is $\{ (a, 0.7),
(b, 0.6), (e, 0.6)\}$, therefore we can say that $S$ is a
possibilistic-${\cal T}$ answer set of $P$. Observe that the
possibilistic-${\cal T}$ answer set semantics and the possibilistic
answer set semantics coincide. In fact, the following proposition
guarantees that both semantics are the same.

\begin{proposition}\label{prop:PossASP-PossFixPointSem}
Let $P$ be a possibilistic disjunctive logic program and $M$ a set
of possibilistic atoms. $M$ is a possibilistic answer set of $P$ if
and only if $M$ is a possibilistic-${\cal T}$ answer set of $P$.
\end{proposition}

\section{Inconsistency in possibilistic logic programs}\label{sec:inconsitency}

In the first part of this section, the relevance of considering inconsistent possibilistic knowledge bases is introduced, and in the second part, some criteria for managing inconsistent possibilistic logic programs are introduced.

\subsection{Relevance of inconsistent possibilistic logic
programs} 

Inconsistent knowledge bases are usually regarded as an
\emph{epistemic hell} that have to be avoided at all costs. However,
many times it is difficult or impossible to stay away from managing
inconsistent knowledge bases. There are authors such as Oct\'avio Bueno  \cite{Bue06} who argues that the consideration of inconsistent systems is a useful device for a number of
reasons: (1) it is often the only way to explore inconsistent
information without arbitrarily rejecting precious data. (2)
inconsistent systems are sometimes the only way to obtain
new information (particularly information that conflicts with deeply
entrenched theories). As a result, (3) inconsistent belief
systems allow us to make better \emph{informed decisions} regarding
which bits of information to accept or reject in the end.

In order to give a small example, in which exploring
inconsistent information can be important for making a better informed
decision, we will continue with the medical scenario described in
Section \ref{sec:PossProg-Motivation}. In Example
\ref{example:reduction}, we have already presented the grounded
program $P_{infections}$ of our medical scenario: \\

{\small \noindent  \textbf{probable}: $r\_inf(present, 1) \vee
no\_r\_inf(present,1)$ $\leftarrow$ $action(transplant, 0),$ \\
$~~~~~~~~~~~~~~~~~~~~~~~~~~~~~~~~~~~~~~~~~~~~~~~ d\_inf(present, 0).$ \\
\noindent \textbf{confirmed}: $o(good\_graft\_funct, 1) \vee
o(delayed\_graft\_funct, 1)\vee$ \\
$~~~~~~~~~~~~~~~~~~~~~~ o(terminal\_insufficient\_funct, 1)
\leftarrow  action(transplant, 0).$ \\
\noindent \textbf{confirmed}: $action(transplant,0) \leftarrow$
$o(terminal\_insufficient\_funct, 0)$.\\
\noindent \textbf{plausible}: $cs(stable, 1) \leftarrow
o(good\_graft\_funct, 1)$.\\
\noindent \textbf{plausible}: $cs(unstable, 1) \leftarrow
o(delayed\_graft\_funct, 1)$.\\
\noindent \textbf{plausible}: $cs($0-urgency$, 1) \leftarrow
o(terminal\_insufficient\_funct, 1),$ \\
$~~~~~~~~~~~~~~~~~~~~~~~~~~~~~~~~~~~~~~~~~~~~~~ action(transplant,0)$.\\
\noindent \textbf{certain}: $\;\; \leftarrow action(transplant, 0),
action(wait, 0)$. \\
\noindent \textbf{certain}: $\;\; \leftarrow action(transplant,0),
cs(dead,0)$. \\
\noindent \textbf{certain}: $ d\_inf(present,0)$. \\
\noindent \textbf{certain}: $ no\_r\_inf(present,0)$. \\
\noindent \textbf{certain}: $ o(terminal\_insufficient\_funct,0)$. \\
\noindent \textbf{certain}: $cs(stable, 0)$. \\
}

\noindent As mentioned in Example \ref{example:reduction}, in this
program the atoms $\lnot r\_inf(present,0)$ and \\$\lnot
r\_inf(present,1)$ were replaced by $no\_r\_inf(present,0)$ and
$no\_r\_inf(present,1)$ respectively. Usually in standard answer set
programming, the constraints \\

\noindent $\leftarrow no\_r\_inf(present,0), r\_inf(present,0).$\\
\noindent $\leftarrow no\_r\_inf(present,1), no\_r\_inf(present,1)$. \\

\noindent must be added to the program to avoid inconsistent
answer sets. In order to illustrate the role of these kinds of
constraints, let $C_1$ be the following possibilistic constraints: \\

\noindent \textbf{certain}: $\;\; \leftarrow no\_r\_inf(present,0), r\_inf(present,0).$\\
\noindent \textbf{certain}: $\;\; \leftarrow no\_r\_inf(present,1),
no\_r\_inf(present,1)$. \\

Also let us consider three new possibilistic clauses (denoted by
$P_{v}$):\\

\noindent \textbf{confirmed}: $v(kidney,0) \leftarrow cs(stable,1), action(transplant,0)$. \\
\noindent \textbf{probable}: $no\_v(kidney,0) \leftarrow r\_inf(present,1), action(transplant,0)$. \\
\noindent \textbf{certain}: $\;\; \leftarrow \; not \;cs(stable,1)$. \\

\noindent The intended meaning of the predicate $v(t,T)$ is that the
organ $t$ is viable for a transplant and $T$ denotes a moment in time. Observe that we replaced the atom $\lnot v(kidney,0)$ with $no\_v(kidney,0)$. The reading of the first clause is that if the clinical situation of the organ recipient is stable after the graft, then it is \emph{confirmed} that the kidney is viable for transplant. The reading of the second one is that if the organ recipient is infected after the graft, then it is \emph{plausible} that the kidney is not viable for transplant.
The aim of the possibilistic constraint is to discard scenarios
in which the clinical situation of the organ recipient is not
stable.  Let us  consider the respective possibilistic
constraint \wrt the atoms $no\_v(kidney,0)$ and $v(kidney,0)$ (denoted
by $C_2$): \\

\noindent \textbf{certain}: $\;\; \leftarrow no\_v(kidney,0), v(kidney,0).$\\

Two programs are defined: $$P := P_{infections} \; \cup \; P_{v}
\text{ and }P_c := P_{infections} \; \cup P_{v} \; \cup \; C_1 \;
\cup \;C_2$$

\noindent Basically, the difference between $P$ and
$P_c$ is that $P$ allows inconsistent possibilistic models and $P_c$
does not allow inconsistent possibilistic models.

Now let us consider the possibilistic answer sets of the programs
$P$ and $P_c$. One can see that the program $P_c$ has just one
possibilistic answer set:\\

\noindent $\{ (d\_inf(present,0), certain),$
$(no\_r\_inf(present,0), certain),$\\
$(o(terminal\_insufficient\_funct,0),certain),$
$(cs(stable,0),certain),$ \\$(action(transplant,0),
confirmed),$ $(o(good\_graft\_funct,1), confirmed),$ \\
$\textbf{(cs(stable,1), plausible)},$
$\textbf{(no\_r\_inf(present,1), probable)},$\\
$\textbf{(v(kidney,0), plausible)}\}$\\

\noindent This possibilistic answer set suggests that since it is
plausible that the recipient's clinical situation will be stable after
the graft, it is plausible that the kidney is \emph{viable} for
transplanting. \emph{Observe that the possibilistic answer sets of
$P$ do not show the possibility that the organ recipient could be infected after the
graft}.

Let us consider the possibilistic answer set of the program $P$: \\

\noindent $S_1 := \{ (d\_inf(present,0), certain),$
$(no\_r\_inf(present,0), certain),$\\
$(o(terminal\_insufficient\_funct,0),certain),$
$(cs(stable,0),certain),$ \\$(action(transplant,0),
confirmed),$ $(o(good\_graft\_funct,1), confirmed),$ \\
$\textbf{(cs(stable,1), plausible)},$
$\textbf{(no\_r\_inf(present,1), probable)},$\\
$\textbf{(v(kidney,0), plausible)}\}$\\

\noindent $S_2 := \{ (d\_inf(present,0), certain),$
$(no\_r\_inf(present,0), certain),$\\
$(o(terminal\_insufficient\_funct,0),certain),$
$(cs(stable,0),certain),$ \\$(action(transplant,0),
confirmed),$ $(o(good\_graft\_funct,1), confirmed),$ \\
$\textbf{(cs(stable,1), plausible)},$
$\textbf{(r\_inf(present,1), probable)},$\\
$\textbf{(v(kidney,0), plausible)}, \textbf{(no\_v(kidney,0), probable)}\}$\\

\noindent $P$ has two possibilistic answer sets: $S_1$ and $S_2$.
$S_1$ corresponds to the possibilistic answer set of the program
$P_c$ and $S_2$ is an inconsistent possibilistic answer set ---
because the atoms $\textbf{(v(kidney,0), plausible)}$ and
$\textbf{(no\_v(kidney,0), probable)}$ appear in $S_2$. Observe that
although $S_2$ is an inconsistent possibilistic answer set, it
contains important information \wrt the considerations of our
scenario. $S_2$ suggests that even though it is plausible that the
clinical situation of the organ recipient will be stable after the
graft, it is also probable that the organ recipient will be infected
by the infection of the donor's organ.

Observe that $P_c$ is unable to infer the possibilistic
answer set $S_2$; because, it contains the following possibilistic constraint:\\

\noindent\noindent \textbf{certain}: $\;\; \leftarrow no\_v(kidney,0), v(kidney,0).$\\

By defining these kinds of constraints, we can guarantee that any
possibilistic answer set inferred from $P_c$ will be consistent;
however, one can omit important considerations \wrt a
decision-making problem. In fact, we agree with Bueno \cite{Bue06}
that considering inconsistent systems as inconsistent possibilistic
answer sets is some times the only way to explore inconsistent
information without arbitrarily rejecting precious data.

\subsection{Inconsistency degrees of possibilistic sets}

To manage inconsistent possibilistic answer sets, it is necessary
to define a criterion of preference between possibilistic answer
sets. In order to define a criterion between possibilistic answer
sets, the concept of \emph{inconsistency degree of a possibilistic set} is defined. We say that a set of possibilistic atoms $S$ is
inconsistent (resp. consistent) if and only if $S^*$ is inconsistent
(resp. consistent), that is to say there is an atom $a$ such that $a,\lnot a
\in S^*$.

\begin{definition}
Let ${\cal A} \in {\cal SP}$.
The inconsistent degree of $S$ is defined as follows:

\[
InconsDegre(S) := \left\{
  \begin{array}{lcl}
\bot_{\cal Q} & \;\;\;\;\; & \text{ if } S^* \text{ is consistent} \\
{\cal GLB}(\{ \alpha | S_{\alpha} \text{ is consistent}  \})&&
\text{otherwise }
  \end{array}
\right.
\]

\noindent in which $\bot_{\cal Q}$ is the bottom of the lattice
(${\cal Q}$,$\leq$) and $S_{\alpha} := \{ (a, \alpha_1) \in S |
\alpha_1 \geq \alpha \}$.
\end{definition}

For instance, the possibilistic answer set $S_2$ of our example
above has a degree of inconsistency of \emph{confirmed}. Based on the degree of
inconsistency of possibilistic sets, we can define a
criterion of preference between possibilistic answer sets.

\begin{definition}
Let  $P = \langle ({\cal Q},\leq), N \rangle$ be a possibilistic
program and $M_1$, $M_2$ two possibilistic answer sets of $P$. We say
that $M_1$ is more-consistent than $M_2$ if and only if
$InconsDegre(M_1) < InconsDegre(M_2)$.
\end{definition}

In our example above, it is obvious that $S_1$ is more-consistent than $S_2$. In general terms, a possibilistic answer set $M_1$ is preferred to $M_2$ if and only
if $M_1$ is more-consistent than $M_2$. This means that any
consistent possibilistic answer set will be preferred to any
inconsistent possibilistic answer set.

So far we have commented only on the case of inconsistent possibilistic
answer sets. However, there are possibilistic programs that are
inconsistent because they have no possibilistic answer sets.
For instance, let us consider the
following possibilistic program $P_{inc}$ (we are assuming the
lattice of Example \ref{example:poss-resolution}):

\begin{tabular}{rl}\\
$0.3: $ & $a \leftarrow \; not \; b.$ \\
$0.5: $ & $b \leftarrow \; not \; c.$ \\
$0.6: $ & $c \leftarrow \; not \; a.$ \\
\end{tabular}\\

\noindent Observe that $P_{inc}^*$ has no answer sets;
hence, $P_{inc}$ has no possibilistic answer sets.

\subsection{Restoring inconsistent possibilistic knowledge bases}

In order to restore consistency of an inconsistent possibilistic
knowledge base, possibilistic logic eliminates the set of possibilistic
formul\ae\ which are lower than the inconsistent degree of the
inconsistent knowledge base. Considering this idea, the authors
of \cite{NicGarSteLef06} defined the concept of $\alpha$-cut for
possibilistic logic programs. Based on Definition 14 of
\cite{NicGarSteLef06}, we define its respective generalization for
our approach.

\begin{definition}
Let $P$ be a possibilistic logic program

\begin{description}
\item[-] the strict $\alpha$-cut is the subprogram $P_{>\alpha} = \{r\in P | n(r) > \alpha \}$

\item[-] the consistency cut degree of $P$:

\[
ConsCutDeg(P) := \left\{
  \begin{array}{lcl}
\bot_{\cal Q} & \;\;\;\;\; & \text{ if } P^* \text{ is consistent} \\
{\cal GLB}(\{ \alpha | P_{\alpha} \text{ is consistent}  \})&&
\text{otherwise }
  \end{array}
\right.
\]
\end{description}

\noindent where $\bot_{\cal Q}$ is the bottom of the lattice
(${\cal Q}$,$\leq$).

\end{definition}

Notice that the consistency cut degree of a possibilistic logic
program identifies the minimum level of certainty for which a strict
$\alpha$-cut of $P$ is consistent. As Nicolas  \etal, remarked in
\cite{NicGarSteLef06}, by the non-monotonicity of the framework, it
is not certain that a higher cut is necessarily consistent.

In order to illustrate these ideas, let us reconsider  the
program $P_{inc}$. First, one can see that $ConsCutDeg(P_{inc})=
0.3$; hence, the subprogram $P_{ConsCutDeg(P_{inc})}$ is:

\begin{tabular}{rl}\\
$0.5: $ & $b \leftarrow \; not \; c.$ \\
$0.6: $ & $c \leftarrow \; not \; a.$ \\
\end{tabular}\\

\noindent Observe that this program has a possibilistic answer set
which is $\{ (c, 0.6)\}$.
Hence due to the strict $\alpha$-cut of $P$, one is able to infer information from $P_{inc}$

To resume, one can identify two kinds of inconsistencies in our approach,

\begin{itemize}
  \item one which arises from the presence of complementary atoms in a
possibilistic answer set  and
  \item one which arises from the non-existence of a possibilistic
answer set of a possibilistic logic program.
\end{itemize}

\noindent To manage the inconsistency of possibilistic answer
sets, a criterion of preference between possibilistic answer sets was defined.
On the other hand, to manage the non-existence of a possibilistic answer set
of a possibilistic logic program $P$, the approach suggested by Nicolas \etal $~$in \cite{NicGarSteLef06}, was adopted. 
This approach is based on $\alpha$\emph{-cuts} in order  to get consistent subprograms of a given program $P$.

%
%
%

\section{Related Work}\label{sec:relatedWorkPossDisjunctivePrograms}

Research on logic programming with uncertainty has dealt with
various approaches of logic programming semantics, as well as
different applications. Most of the approaches in the literature
employ one of the following formalisms:

\begin{itemize}
  \item annotated logic programming, \eg $~$\cite{KifSub92}.
  \item probabilistic logic, \eg $~$ \cite{NgS92,Lukasiewicz98,Kern-IsbernerL04,BaralGR09}.
  \item fuzzy set theory, \eg $~$
  \cite{Emden86,RodRome08,NieuwenborghCV07}.
  \item multi-valued logic, \eg $~$ \cite{Fitting91,Lakshmanan94}.
  \item evidence theoretic logic programming, \eg $~$ \cite{Bal07}.
  \item possibilistic logic, \eg $~$
  \cite{DubLanPra01,AlsinetG02,AlsGod00,AlsinetCGS08,NicGarSteLef06}.
\end{itemize}

Basically, these approaches differ in the underlying notion of uncertainty and how uncertainty values, associated with clauses and facts, are managed. Among these approaches, the formalisms based on possibilistic logic are closely related to the approach presented in this paper. A clear distinction betweem them and the formalism of this paper is that none of them capture disjunctive clauses. On the other hand, excepting the work of Nicolas, \etal, \cite{NicGarSteLef06}, none of these approaches describe a formalism for dealing with uncertainty in a logic program with default negation by means of possibilistic logic. Let us recall that the  work of \cite{NicGarSteLef06} is totally captured by the formalism presented in this paper (Proposition \ref{prop:PosstASP-PossStable}), but not directly vice versa. For instance, let us consider the possibilistic logic programs $P = \langle (\{0.1, \dots, 0.9\},\leq), N \rangle$ such that $\leq$ is the standard relation  between rational number and $N$ the following set of possibilistic clauses:\\

\begin{tabular}{lll}
$0.5: a \vee b.$ & $~~~~$ & $0.5: a \leftarrow b.$ \\
                 & $~~~~$ & $0.5: b \leftarrow a.$ \\
\end{tabular}\\

\noindent By considering a standard transformation from disjunctive clauses to normal clauses \cite{Bar03}, this program can be transformed to the possibilistic normal logic programs $P'$:\\

\begin{tabular}{lll}
$0.5: a \leftarrow \; not \; b.$ & $~~~~$ & $0.5: a \leftarrow b.$ \\
$0.5: b \leftarrow \; not \; a.$ & $~~~~$ & $0.5: b \leftarrow a.$ \\
\end{tabular}\\

\noindent One can see that $P$ has a possibilistic answer set: $\{(a, 0.5), (b, 0.5)\}$; however, $P'$ has no possibilistic answer sets.

Even though, one can find a wide range of formalisms for dealing with uncertainty by using \emph{normal logic programs}, there are few proposals for dealing with uncertainty by using \emph{disjunctive logic programs} \cite{Lukasiewicz01,GerRonPan01,Mateis00,BaralGR09}:

\begin{itemize}
  \item In \cite{Lukasiewicz01}, Many-Valued Disjunctive Logic Programs with \emph{probabilistic
semantics} are introduced. In this approach, \emph{probabilistic values} are associated with each clause. Like our approach, Lukasiewicz considers partial evaluation for characterizing different semantics by means of probabilistic theory.
  \item In \cite{GerRonPan01}, the logic programming language \emph{Disjunctive Chronolog} is introduced. This approach combines \emph{temporal and disjunctive logic programming}. Disjunctive Chronolog is capable of expressing dynamic behaviour as well as uncertainty. In this approach, like our semantics, it is shown that logic semantics of these programs can be characterized by a fixed- point semantics.
  \item In \cite{Mateis00}, the Quantitative Disjunctive Logic Programs (QDLP) are introduced. These programs associate an reliability interval with each clause. Different triangular norms (T-norms) are employed to define calculi for propagating uncertainty information from the premises to the conclusion of a quantitative rule; hence, the semantics of these programs is parameterized. This means that each choice of a T-norm induces different QDLP languages.
  \item In \cite{BaralGR09}, intensive research is done in order to achieve a complete integration between ASP and probability theory. This approach is similar to the approach presented in this paper; but it is in the context of probabilistic theory.
\end{itemize}

We want to point out that the syntactic approach of this paper is motivated by the fact that the possibilistic logic is axiomatizable; therefore, a proof theory approach (axioms and inference rules) leads to constructions of a possibilistic semantics such as a logic inference. This kind of possibilistic framework allows us to explore extensions of the possibilistic answer set semantics by considering the inference of different logics. In fact, by considering a syntactic approach, one can explore properties such as \emph{strong equivalence} and \emph{free-syntax programs}. This means that the exploration of a syntactic approach leads to important implications such as the implications of an approach based on interpretations and possibilistic distributions.

The consideration of axiomatizations of given logics has shown to be a generic approach for characterizing logic programming semantics. For instance, the answer set semantics inference can be characterized as \emph{a logic inference} in terms of the proof theory of intuitionistic logic and intermediate logics \cite{Pea99,OsNaAr:tplp}.

\section{Conclusions and future work}

At the beginning of this research, two main goals were expected to be achieved: 1.- a possibilistic extension of the answer set programming paradigm for leading with uncertain, inconsistent and incomplete information; and,  2.- exploring the axiomatization of possibilistic logic in order to define a computable possibilistic disjunctive semantics.

In order to achieve the first goal, the work presented in \cite{NicGarSteLef06} was taken as a reference point. Unlike the approach of \cite{NicGarSteLef06}, which is restricted to \emph{possibilistic normal programs}, we define a possibilistic logic programming framework based on \emph{possibilistic disjunctive logic programs}. Our approach introduces the use of
possibilistic disjunctive clauses which are able to capture \emph{incomplete information} and \emph{incomplete states of a knowledge base} at the same time.

For capturing the semantics of possibilistic disjunctive logic programs, the axiomatization of possibilistic logic and the standard definition of the answer set semantics are taken as a base. Given that the inference of possibilistic logic is characterized by a possibilistic resolution rule (Proposition \ref{prop:poss-R-sound}), it is shown that:
\begin{enumerate}
  \item The optimal certainty value of an atom which belongs to a possibilistic answer set corresponds to the optimal refutation by possibilistic resolution (Proposition \ref{prop:poss-R-Complete}); hence,
  \item There exists an algorithm for computing the possibilistic answer sets of a possibilistic disjunctive logic program (Proposition \ref{prop:algorithm-PossASP}).
\end{enumerate}

As an alternative approach for inferring the possibilistic answer set semantics, it is shown that this semantics can be characterized by a simplified version of \emph{the principle of partial evaluation}. This means that the possibilistic answer set semantics is characterized by a possibilistic fixed-point operator (Proposition \ref{prop:PossASP-PossFixPointSem}). This result gives two points of view for constructing the possibilistic answer set semantics in terms of two \emph{syntactic processes} (\ie, the possibilistic proof theory and the principle of partial evaluation).

Based on the flexibility of possibilistic logic for defining degrees of uncertainty, it is shown that non-numerical degrees for capturing uncertain information can be captured by the defined possibilistic answer set semantics. This is illustrated in a medical scenario.

To manage the inconsistency of possibilistic models, we have defined a criterion of preference between possibilistic answer sets. Also, to manage the non-existence of possibilistic answer set of a possibilistic logic program $P$, we have adopted the approach suggested by Nicolas \etal $~$ in \cite{NicGarSteLef06} of cuts for achieving consistent subprograms of $P$.

In future work, there are several topics which will be explored. One of the main topics to explore is to show that the possibilistic answer set semantics can be characterized as a logic inference in terms of a possibilistic version of the intuitionistic logic. This issue is motivated by the fact that the answer set semantic inference can be characterized as a logic inference in terms of intuitionistic logic \cite{Pea99,OsNaAr:tplp}.
On the other hand, we have been exploring to define a possibilistic action language. In \cite{NOCCL07}, we have already defined our first ideas in the context of the action language ${\cal A}$. Finally, we have started to explore the definition of a possibilistic framework in order to define preference between rules and preferences between atoms. With this objective in mind, possibilistic ordered disjunction programs have been explored \cite{ConfalonieriNOV09}.

\bibliographystyle{acmtrans}
\bibliography{../../biblio/papers_jcns}

\section*{Appendix: Proofs}\label{appendix:ProofPossPrograms}

In this appendix, we give the proofs of some results presented in this paper.

\begin{statement}[Proposition \ref{prop:ASPPossInference-LUB}]
Let $P =  \langle ({\cal Q},\leq), N \rangle$ be a possibilistic disjunctive logic program and $M_1,M_2 \in {\cal PS}$ such that $M_1^* = M_2^*$. If $P \Vvdash_{PL} M1$ and $P \Vvdash_{PL} M2$, then $P \Vvdash_{PL} M_1 \sqcup M_2$.
\end{statement}
\begin{proof}
The proof is straightforward.
\end{proof}

\begin{statement}[Proposition \ref{prop:PostASP-ASP}]
Let $P$ be a possibilistic disjunctive logic program. If $M$ is a
possibilistic answer set of P then $M^*$ is an answer set of $P^*$.
\end{statement}
\begin{proof}
The proof is straightforward by the possibilistic answer set's
definition.
\end{proof}

\begin{statement}[Proposition \ref{prop:Post-top-ASP-ASP}]
Let $P = \langle ({\cal Q},\leq), N \rangle$ be a possibilistic disjunctive logic program and
$\alpha$ be a fixed element of ${\cal Q}$.
If $\forall r \in P$, $n(r) = \alpha$  and $M'$ is an answer set of $P^*$, then $M :=\{
(a, \alpha)| a \in M' \}$ is a possibilistic answer set of $P$.
\end{statement}
\begin{proof}
Let us introduce two observations:

\begin{enumerate}
  \item We known that $\forall r \in P$, $n(r) = \alpha$; hence, if $P \vdash_{PL} (a, \alpha')$, then $\alpha' = \alpha$. This statement can be proved by contradiction.
  \item Given a set of atoms $S$, if $\forall r \in P$, $n(r) = \alpha$ then $\forall r \in P_S$, $n(r) = \alpha$. This statement follows by fact that the reduction $P_S$ (Definition \ref{def:reduction}) does not affect $n(r)$ of clause rule $r$ in $P$.
\end{enumerate}

As premises we know that  $\forall r \in P$, $n(r) = \alpha$  and $M'$ is an answer set of $P^*$; hence, let $M :=\{ (a, \alpha)| a \in M' \}$. Hence, we will prove that $M$ is a possibilistic answer set of $P$.

If $M'$ is an answer set of $P^*$, then $\forall a \in M'$, $\exists \alpha' \in {\cal Q}$ such that $P_{M'} \vdash_{PL} (a, \alpha')$. Therefore, by observations 1 and 2, $\forall a \in M'$,  $P_{M'} \vdash_{PL} (a, \alpha)$. Then, $P \Vvdash_{PL} M$. Observe that $M$ is a greatest set in ${\cal PS}$, hence, since $P \Vvdash_{PL} M$ and $M$ is a greatest set, $M$ is a possibilistic answer set of $P$.

%
\end{proof}


\begin{statement}[Proposition \ref{prop:PosstASP-PossStable}]
Let $P := \langle (Q,\leq), N \rangle$ be a possibilistic normal
program such that $(Q,\leq)$ is a totally ordered set and ${\cal
L}_{P}$ has no extended atoms. $M$ is a possibilistic answer set of
P if and only if $M$ is a possibilistic stable model of $P$.
\end{statement}
\begin{proof}(Sketch)
It is not difficult to see that when $P$ is a possibilistic normal
program, then the syntactic reduction of Definition
\ref{def:reduction} and the syntactic reduction of Definition 10
from \cite{NicGarSteLef06} coincide. Then the proof is reduced to
possibilistic definite programs. But, this case is straightforward,
since essentially GMP is applied for inferring the possibilistic
models of the program in both approaches.
\end{proof}

\begin{statement}[Proposition \ref{prop:poss-R-sound}]
Let ${\cal C}$ be a set of possibilistic disjunctions, and $C = (c
\; \alpha)$ be a possibilistic clause obtained by a finite number of
successive application of \emph{(R)} to ${\cal C}$; then ${\cal C}
\vdash_{PL} C$.
\end{statement}
\begin{proof}(The proof is similar to the proof of Proposition 3.8.2 of
\cite{Dubois94}) Let us consider two possibilistic clauses:  $C_1 =
(c_1 \; \alpha_1)$ and $C_2 =(c_2 \; \alpha_2)$, the application of
$R$ yields $C' = (R(c_1,c_2) \; {\cal GLB}(\{\alpha_1,
\alpha_2\}))$. By classic logic, we known that $R(c_1,c_2)$ is
sound; hence the key point of the proof is to show that
$n(R(c_1,c_2)) \geq {\cal GLB}(\{\alpha_1, \alpha_2\})$.

By definition of necessity-valued clause, $n(c_1) \geq \alpha_1$ and
$n(c_2)\geq \alpha_2$, then  $n(c_1 \wedge c_2 ) = {\cal GLB}(\{
n(c_1), n(c_2) \}) \geq {\cal GLB}(\{ \alpha_1 , \alpha_2 \})$.
Since $c_1 \wedge c_2 \vdash_{C} R(c_1, c_2)$, then $n(R(c_1, c_2))
\geq n(c_1 \wedge c_2)$ (because if $\varphi \vdash_{PL} \psi$ then
$N(\psi)\geq N(\varphi)$). Thus $n(R(c_1, c_2)) \geq {\cal GLB}(\{
\alpha_1 , \alpha_2 \})$; therefore (R) is sound. Then by induction
any possibilistic formula inferred by a finite number of successive
applications of (R) to ${\cal C}$ is a logical consequence of ${\cal
C}$.
\end{proof}

\begin{statement}[Proposition \ref{prop:poss-R-Complete}]
Let $P$ be a set of possibilistic clauses and ${\cal C}$ be the set
of possibilistic disjunctions obtained from $P$; then the valuation
of the optimal refutation by resolution from ${\cal C}$  is the
inconsistent degree of $P$.
\end{statement}
\begin{proof}(The proof is similar to the proof of Proposition 3.8.3 of \cite{Dubois94})
By possibilistic logic, we know that ${\cal C} \vdash_{PL} (\bot \;
\alpha)$ if and only if $({\cal C}_{\alpha})^*$ is inconsistent in
the sense of classic logic. Since (R) is complete in classic logic,
then there exists a refutation $R(\square)$ from $({\cal
C}_{\alpha})^*$. Thus considering the valuation of the refutation
$R(\square)$, we obtain a refutation from ${\cal C}_{\alpha}$ such
that $n(R(\square)) \geq \alpha$. Then  $n(R(\square)) \geq
Inc({\cal C})$. Since (R) is sound then $n(R(\square))$ cannot be
strictly greater than $Inc({\cal C})$. Thus $n(R(\square))$ is equal
to $Inc({\cal C})$. According to Proposition 3.8.1 of
\cite{Dubois94}, $Inc({\cal C}) = Inc (P)$, thus $n(R(\square))$ is
also equal to $Inc(P)$.
\end{proof}







\begin{statement}[Proposition \ref{prop:algorithm-PossASP}]
Let $P := \langle ({\cal Q},\leq), N \rangle$ be a possibilistic
logic program. The set Poss-ASP returned by $Poss\_Answer\_Sets(P)$
is the set of all the possibilistic answer sets of $P$.
\end{statement}
\begin{proof}
The result follows from the following facts:

\begin{enumerate}
  \item The function $ASP$ computes all the answer set of $P^*$.
  \item If $M$ is a possibilistic answer set of P iff
  $M^*$ is an answer set of $P^*$ (Proposition
  \ref{prop:PostASP-ASP}).
  \item By Corollary \ref{corrollary:optimal-PossValue}, we know
  that the possibilistic resolution rule $R$ is sound and complete
  for computing optimal possibilistic degrees.
\end{enumerate}
\end{proof}

\begin{statement}[Proposition \ref{prop:FixPointOperatorPossSem}]
Let $P$ be a possibilistic disjunctive logic program. If $\Gamma_0
:= {\cal T}(P)$ and $\Gamma_i := {\cal T}(\Gamma_{i-1})$ such that
$i \in {\cal N}$, then $ \exists \; n \in {\cal N}$ such that
$\Gamma_n = \Gamma_{n-1}$. We denote $\Gamma_n$ by $\Pi(P)$.
\end{statement}
\begin{proof}
It is not difficult to see that the operator ${\cal T}$ is
monotonic, then the proof is direct by Tarski's Lattice-Theoretical
Fixpoint Theorem \cite{Tar55}.
\end{proof}

\begin{statement}[Proposition
\ref{prop:PossASP-PossFixPointSem}] Let $P$ be a possibilistic
disjunctive logic program and $M$ a set of possibilistic atoms. $M$
is a possibilistic answer set of $P$ if and only if $M$ is a
possibilistic-${\cal T}$ answer set of $P$.
\end{statement}
\begin{proof}
Two observations:
\begin{enumerate}
\item By definition, it is straightforward that if $M_1$ is a possibilistic answer set of
$P$, then there exists a  possibilistic-${\cal T}$ answer set $M_2$
of $P$ such that $M_1^* = M_2^*$ and viceversa.

\item Since G-GPPE can be regarded as a macro of the possibilistic
rule $(R)$, we can conclude by Proposition \ref{prop:poss-R-sound}
that G-GPPE is sound.

\end{enumerate}

Let $M_1$ be a possibilistic answer set of $P$ and $M_2$ be a
possibilistic-${\cal T}$ answer set of $P$. By Observation 1, the
central point of the proof is to prove that if $(a, \alpha_1) \in
M_1$ and $(a, \alpha_2) \in M_2$ such that $M_1^* = M_2^*$, then
$\alpha_1 = \alpha_2$.

The proof is by contradiction. Let us suppose that $(a, \alpha_1)
\in M_1$ and $(a, \alpha_2) \in M_2$ such that $M_1^* = M_2^*$ and
$\alpha_1 \neq \alpha_2$. Then there are two cases $\alpha_1 <
\alpha_2$ or $\alpha_1 > \alpha_2$

\begin{description}
\item[$\alpha_1 < \alpha_2$]: Since G-GPPE is sound (Observation 2),
then $\alpha_1$ is not the optimal necessity-value for the atom $a$,
but this is false by Corollary \ref{corrollary:optimal-PossValue}.

\item[$\alpha_1 > \alpha_2$]: If $\alpha_1 > \alpha_2$ then there
exists a possibilistic claus $\alpha_1 : {\cal A} \leftarrow {\cal
B}^+ \in P^{(M_1)^*}$ that belongs to the optimal refutation of the
atom $a$ and it was not reduced by G-GPPE. But this is false because
G-GPPE is a macro of the resolution rule (R).
\end{description}

\end{proof}

\end{document}